\documentclass{article}
\usepackage{log_2023}            % for camera-ready version
\newif\ifnbld
\newif\ifconf
\newif\iftr
\newif\ifEXT
\EXTfalse

\trtrue
\conffalse

\trfalse
\conftrue

\nbldfalse
\nbldtrue

\usepackage[numbers, compress, sort]{natbib}
\newcommand{\nameA}{HOT}
\newcommand{\nameAS}{HOT\ }

\usepackage[utf8]{inputenc} % allow utf-8 input
\usepackage[T1]{fontenc}    % use 8-bit T1 fonts
\usepackage{url}            % simple URL typesetting
\usepackage{booktabs}       % professional-quality tables
\usepackage{amsfonts}       % blackboard math symbols
\usepackage{nicefrac}       % compact symbols for 1/2, etc.
\usepackage{microtype}      % microtypography
\usepackage{xcolor}         % colors

\usepackage{wrapfig}

\newif\ifsq     % Squeeze space?
\newif\ifsqCAP
\newif\ifsqVS
\newif\ifsqEN
\newif\ifsqTIT

\newcommand{\ignore}[1]{}

\sqtrue
\sqCAPfalse
\sqENtrue
\sqVStrue
\sqTITtrue

\usepackage{etex}
\usepackage{balance}
\usepackage{epstopdf}
\usepackage{placeins}

\usepackage{graphicx}
\usepackage{float}
\usepackage{dblfloatfix}
\usepackage{multirow}
\usepackage{rotating}
\usepackage{makecell}
\usepackage{tabulary}
\usepackage{parcolumns}
\usepackage{tikz}
\usetikzlibrary{tikzmark}

\usepackage{xpatch}
\expandafter\xpatchcmd
\csname pgfk@/tikz/every picture/.@cmd\endcsname
{\thepage}{\arabic{page}}{}{}

\tikzstyle{comment} = [draw, fill=blue!70, text=white, text width=3cm, minimum height=1cm, rounded corners, align=left, font=\scriptsize]
\tikzstyle{background_alg} = [draw, fill=blue!20, opacity=0.4, inner sep=4pt, rounded corners=2pt]

\usetikzlibrary{shapes}
\usetikzlibrary{plotmarks}
\usetikzlibrary{calc, fit}

\usepackage{enumitem}

\usepackage{amsthm}
\usepackage{amsmath,amssymb,amsfonts}
\usepackage{mathtools,mathrsfs}

\usepackage{soul}
\usepackage{fontawesome}
\usepackage{pifont}
\usepackage{textcomp}
\usepackage{booktabs}
\usepackage{url}
\usepackage{pbox}
\usepackage[normalem]{ulem}
\usepackage[10pt]{moresize}
\ifsqCAP
\usepackage[font={normalfont, scriptsize}]{caption}
\usepackage[font={normalfont, scriptsize}]{subcaption}
\else
\usepackage[font={normalfont, scriptsize}]{caption}
\usepackage[font={normalfont, scriptsize}]{subcaption}
\fi

\newcommand{\vspaceSQ}[1]{\ifsqVS\vspace{#1}\fi}
\newcommand{\enlargeSQ}[1]{\ifsqEN\enlargethispage{\baselineskip}\fi}

\ifsqTIT
\usepackage[compact]{titlesec}
\titlespacing*{\section}{0pt}{3pt}{-1pt}
\titlespacing*{\subsection}{0pt}{0pt}{-3pt}
\titlespacing*{\subsubsection}{0pt}{2pt}{1pt}
\fi

\usepackage{xcolor}
\definecolor{darkgrey}{RGB}{70,70,70}
\definecolor{lightgrey}{RGB}{200,200,200}
\definecolor{lyellow}{RGB}{255,255,100}
\definecolor{llyellow}{RGB}{250,250,180}
\definecolor{lgreen}{RGB}{144,238,144}
\definecolor{raphael_comments}{RGB}{13, 145, 24}

\usepackage[customcolors]{hf-tikz}
\hfsetbordercolor{white}
\hfsetfillcolor{vlgray}

\definecolor{vlgray}{rgb}{0.77 0.77 0.77}
\definecolor{ablack}{rgb}{0.2 0.2 0.2}
\definecolor{vllgray}{rgb}{0.9 0.9 0.9}
\definecolor{bblue}{rgb}{0.7 0.7 0.99}

\usepackage{colortbl}

\usepackage{inconsolata}
\usepackage{listings}

\ifsq
\else
\fi

\newcommand{\afonso}[1]{\textcolor{teal}{[Afonso: #1]}}

\definecolor{hlL}{rgb}{0.8 0.8 0.99}

\newcounter{highlight}

\newcounter{hlLR}

\newcounter{hlLIR}

\newcounter{hlLIIR}

\newcounter{Ahighlight}

\usepackage{scalerel,stackengine}
\stackMath
\newcommand\rwh[1]{%
\savestack{\tmpbox}{\stretchto{%
  \scaleto{%
        \scalerel*[\widthof{\ensuremath{#1}}]{\kern-.6pt\bigwedge\kern-.6pt}%
                  {\rule[-\textheight/2]{1ex}{\textheight}}%WIDTH-LIMITED BIG WEDGE
                              }{\textheight}% 
}{0.5ex}}%
\stackon[1pt]{#1}{\tmpbox}%
}

\usepackage[hang,flushmargin]{footmisc}

\renewcommand{\epsilon}{\ensuremath\varepsilon}

\renewcommand{\phi}{\ensuremath{\varphi}}

\if 0

\usepackage[linesnumbered,ruled]{algorithm2e}
\usepackage{multicol}
\SetKwComment{Comm}{$\triangleright$\ }{}
\SetAlFnt{\scriptsize}
\SetAlCapFnt{\scriptsize}
\SetAlCapNameFnt{\scriptsize}
\SetKwInOut{Input}{Input}
\SetKwInOut{Output}{Output}

\makeatletter
\NewDocumentCommand{\LeftComment}{s m}{%
\Statex \IfBooleanF{#1}{\hspace*{\ALG@thistlm}}\(\triangleright\) #2}
\makeatother

\fi

\usepackage{stmaryrd}

\ifsq

\else

\fi

\definecolor{lightgray1}{gray}{0.6}
\definecolor{lightgray2}{gray}{0.8}

\renewcommand{\marginpar}[1]{}
\renewcommand{\hl}[1]{#1}

\title{\nameA: Higher-Order Dynamic Graph Representation Learning with Efficient Transformers}

\ifnbld

\author[M. Besta et al.]{%
Maciej Besta$^{1}$\thanks{First-author contribution} \quad Afonso Claudino Catarino$^{1}$\footnotemark[1] \quad Lukas Gianinazzi$^1$ \quad Nils Blach$^1$\\ \textbf{Piotr Nyczyk}$^2$ \quad \textbf{Hubert Niewiadomski}$^2$ \quad \textbf{Torsten Hoefler}$^1$\\\\
       $^1$Department of Computer Science, ETH Zurich; $^2$Cledar\\
}
\fi

\if 0
\author[M. Besta et al.]{%
Maciej Besta$^{1 \dagger}$ \quad Afonso Claudino Catarino$^{1 \dagger}$ \quad Lukas Gianinazzi$^1$ \quad Nils Blach$^1$\\ \textbf{Piotr Nyczyk}$^2$ \quad \textbf{Hubert Niewiadomski}$^2$ \quad \textbf{Torsten Hoefler}$^1$\\\\
       $^1$Department of Computer Science, ETH Zurich; $^2$Cledar\\
 $^{\dagger}$Alphabetical order\\ %\texttt{\{maciej.besta, torsten.hoefler\}@inf.ethz.ch}
}
\fi

\begin{document}

\maketitle

%\vspaceSQ{-1em}
\begin{abstract}
%\vspaceSQ{-1em}
%
Many graph representation learning (GRL) problems are dynamic, with millions of edges added or removed per second. A fundamental workload in this setting is dynamic link prediction: using a history of graph updates to predict whether a given pair of vertices will become connected. Recent schemes for link prediction in such dynamic settings employ Transformers, modeling individual graph updates as single tokens. In this work, we propose \nameA: a model that enhances this line of works by harnessing higher-order (HO) graph structures; specifically, $k$-hop neighbors and more general subgraphs containing a given pair of vertices. Harnessing such HO structures by encoding them into the attention matrix of the underlying Transformer results in higher accuracy of link prediction outcomes, but at the expense of increased memory pressure. To alleviate this, we resort to a recent class of schemes that impose hierarchy on the attention matrix, significantly reducing memory footprint. The final design offers a sweetspot between high accuracy and low memory utilization. \nameAS outperforms other dynamic GRL schemes, for example achieving 9\%, 7\%, and 15\% higher accuracy than -- respectively -- DyGFormer, TGN, and GraphMixer, for the MOOC dataset. Our design can be seamlessly extended towards other dynamic GRL workloads.
\end{abstract}

\vspaceSQ{-0.5em}
\section{Introduction}
\label{sec:intro}

\enlargeSQ

Analyzing graphs in a dynamic setting, where edges and vertices can be arbitrarily modified, has become an important task. 
\iftr
Such a setting is omnipresent in the computing landscape: new friendships appear~\cite{jodie, socNet2, socNet3}, novel associations between items and clients are created~\cite{ui1, ui2, ui3, ui4, ui5}, or protein interactions change. 
\fi
For example, the Twitter social network may experience even 500 million new tweets in a single day, while retail transaction graphs consisting of billions of transactions are generated every year~\cite{ammar2016techniques}. Other domains where dynamic networks are often used are transportation~\cite{tn1, tn2, tn3, tn4, tn5}, physical systems~\cite{ps1, sanchez2020learning, ps3}, scientific collaboration~\cite{kazemi, skarding2021foundations, xue2022dynamic, besta2021graphminesuite, besta2022probgraph}, and others~\cite{jodie, socNet2, socNet3, ui1, ui2, ui3, ui4, ui5}.
A fundamental task in such a dynamic graph setting is predicting links that would appear in the future, based on the history of previous graph modifications. This task is crucial for understanding the future of the graph datasets, which enables more accurate graph analytics in real-time in production settings such as recommendation systems in online stores. Moreover, it enables better performance decisions, for example by adjusting load balancing strategies with the knowledge of future events~\cite{besta2023thegdi, besta2019practice, sakr2020future}.

Recent years brought intense developments into harnessing graph representation learning (GRL) techniques for the above-described tasks~\cite{kazemi}, resulting in a broad domain called dynamic graph representation learning (DGRL)~\cite{kazemi2020representation, barros2021survey, skarding2021foundations, sizemore2018dynamic, xue2022dynamic, han2021dynamic}. Initially, various approaches have been proposed based on Temporal Random Walks~\cite{cawn, con10}, Sequential Models~\cite{tcl, graphmixer}, Memory Networks~\cite{mn1, mn2}, or the Dynamic Graph Neural Networks (GNNs) paradigm, in which node embeddings are iteratively updated based on the information passed by their neighbors~\cite{dyrep, tgat, tgn, con5, con6, con9}, while considering temporal data from the past~\cite{tgat, tgn}. However, the most recent and powerful schemes such as DyGFormer~\cite{dygformer} instead directly harness the Transformer model~\cite{vaswani2017attention} for the DGRL tasks. The intuition is that dynamic graphs can be modeled as a sequence of updates over time~\cite{dygformer}, and they could hence benefit from Transformers by treating these updates as individual tokens. This enhanced predictions for dynamic graphs~\cite{dygformer}.

In parallel to the developments in DGRL, many GRL schemes have harnessed higher-order (HO) graph structure~\cite{bick2021higher, xu2018powerful, bick2021higher, battiston2020networks, torres2021and, frasca2022understanding, frasca2020sign, liu2019higher, abu2019mixhop}. Fundamentally, they harness relationships between vertices that go beyond simple edges, for example triangles.
\iftr
Other example HO structures are dense subgraphs such as cliques, or paths and trees that connect subgraphs of vertices relevant to a given task. 
\fi
Incorporating HO graph structures results in fundamentally more powerful predictions for many workloads~\cite{morris2019weisfeiler}. This line of works, however, has primarily focused on static GRL.

In this work, we embrace the HO graph structures for higher accuracy in DGRL (\textbf{contribution \#1}). For this, we harness two powerful HO structures: $k$-hop neighbors 
\iftr
(i.e., neighbors at a distance of $k$ hops) 
\fi
and more general subgraphs containing a given pair of vertices. We harness these structures by encoding them appropriately into the attention matrix of the underlying Transformer. This results in higher accuracy of link prediction.
However, harnessing HO leads to substantially larger memory utilization, which limits its applicability. This is because, intuitively, to make a prediction in the temporal setting, one needs to consider the history of past updates. As HO increases the number of graph elements to be considered in such a history, the number of entities to be used is substantially inflated. For example, when relying on the Transformer architecture (as in DyGFormer), the number of tokens that must be used is significantly increased when incorporating HO. For example, when incorporating $x$ 2-hop neighbors for each vertex, the token count increases by a factor of $x$. This becomes even higher for $k > 2$ and more complex HO structures such as triangles. Hence, one needs to decrease the considered length of the history of updates to make the computation feasible. But then, considering shorter history entails lower accuracy, effectively annihilating benefits gained from using HO in the first place.

We tackle this by harnessing the state of the art outcomes from the world of Transformers (\textbf{contribution \#2}), where \emph{one imposes hierarchy into the traditionally ``flat'' attention matrix}. Examples of such recent \emph{hierarchical Transformer models} are RWKV~\cite{peng2023rwkv}, Swin Transformer~\cite{liu2021swin}, hierarchical BERT~\cite{pappagari2019hierarchical}, Nested Hierarchical Transformer~\cite{zhang2022nested}, Hift~\cite{cao2021hift}, or Block-Recurrent Transformer~\cite{brt}. We employ these designs to alleviate the memory requirements of the attention matrix by dividing it into parts, computing attention locally within each part, and then using such blocks to obtain the final outcomes. For concreteness, we pick Block-Recurrent Transformer~\cite{brt}, but our approach can be used with others.
Our final outcome, a model called \nameA, supported by a theoretical analysis (\textbf{contribution \#3}), ensures long-range and high-accuracy dynamic link prediction. It outperforms all other dynamic GRL schemes (\textbf{contribution \#4}), for example achieving 9\%, 7\%, and 15\% higher accuracy than -- respectively -- DyGFormer, TGN, and GraphMixer, on the MOOC dataset. Our work illustrates the importance of HO structures in the temporal dimension.

\iftr
To summarize, we provide the following contributions:

\begin{itemize}
    \item We design \nameA, a model that harnesses the higher-order graph structures for more accurate predictions in dynamic graph representation learning.
    \item We employ hierarchical Transformers as a mechanism to alleviate memory requirements from the higher-order structures in \nameA, picking Block-Recurrent Transformer~\cite{brt} for concreteness.
    \item We provide a theoretical analysis of the \nameAS final design, formally illustrating its tradeoff between memory requirements and compute costs.
    \item We evaluate \nameAS and illustrate that it outperforms a wide selection of other baselines, including JODIE~\cite{jodie}, DyRep~\cite{dyrep}, TGAT~\cite{tgat}, TGN~\cite{tgn}, CAWN~\cite{cawn}, TCL~\cite{tcl}, GraphMixer~\cite{graphmixer}, and DyGFormer~\cite{dygformer}.
\end{itemize}
\fi

\if 0

\paragraph{Groupings}

Rep \cite{kazemi}

Social Networks \cite{jodie, socNet2, socNet3}

User-Item \cite{ui1, ui2, ui3, ui4, ui5}

Traffic Networks \cite{tn1, tn2, tn3, tn4, tn5}

Physical Systems \cite{ps1, ps2, ps3}

Scholars \cite{kazemi, skarding2021foundations, xue2022dynamic}

Discrete-time \cite{dis1, dis2, dis3, dis4, dis5}

Continuous-time \cite{jodie, dyrep, tgat, tgn, con5, con6, tcl, cawn, con9, con10, con11, graphmixer}

Graph Convolution \cite{dyrep, tgat, tgn, con5, con6, con9}

Temporal Random Walks \cite{cawn, con10}

Sequential Models \cite{tcl, graphmixer}

Memory Networks \cite{mn1, mn2}

Computational cost \cite{jodie, dyrep, tgn, con5, con9, con11}

Staleness problem \cite{kazemi}

RNN \cite{r1, tcl}

Libraries \cite{lib1, lib2, lib3}

\fi

    %\vspaceSQ{-0.5em}
\section{Background}
\vspaceSQ{-0.3em}
\label{sec:back}

\iftr
We first introduce fundamental concepts and notation.
\fi

\enlargeSQ

\iftr
\subsection{Graph Model and Representation}

\else

\fi
Static graphs are usually modeled as a tuple $G=(V, E, f, w)$, where $V$ is the set of nodes, $E \subseteq V \times V$ is the set of edges, $f : V \rightarrow \mathbb{R}^{d_N}$ are the node features and $w : E \rightarrow \mathbb{R}^{d_E}$ are the edge features. 
\textbf{Dynamic graphs} are more complex to represent, as it is necessary to capture their evolution over time. In this work, we focus on the commonly used \emph{Continuous-Time Dynamic Graph (CTDG)} representation~\cite{jodie, dyrep, tgat, tgn, con5, con6, tcl, cawn, con9, con10, con11, graphmixer}. A CTDG is a tuple ($G^{(0)}$, $T$), where $G^{(0)}$ $=$ ($V^{(0)}$, $E^{(0)}$, $f^{(0)}$, $w^{(0)}$) represents the initial state of the graph and $T$ is a set of tuples of the form (\textit{timestamp}, \textit{event}) representing events to be applied to the graph at given timestamps. These events could be node additions/deletions, edge additions/deletions, or feature updates.

\iftr
\subsection{Dynamic Graph Tasks and Workloads}

\else

\fi
Assume a CTDG ($G^{(0)}$, $T$), a timestamp $t \in \mathbb{N}$, and an edge ($u$, $v$). In \textbf{dynamic link prediction}, the task is to decide, whether there is some event $e$ pertaining to ($u$, $v$), such that ($t$, $e$) $\in$ \{ ($t'$, $e$) $\in T$ $|$ $t' = t$ \}, while only considering the CTDG ($G^{(0)}$, \{ ($t'$, $e$) $\in T$ $|$ $t' < t$ \}).
This problem is usually considered in two settings, the \emph{transductive} setting (predicting links between nodes that were seen during training) and the \emph{inductive} setting (predicting links between  nodes not seen during training).
%
\if 0
The second major DGRL task is \emph{dynamic node classification}. Here, we are given a CTDG ($G^{(0)}$, $T$), a timestamp $t \in \mathbb{N}$, a node $v \in V$, and a set of $N \in \mathbb{N}$ classes $\{S_i\}_{i=1}^N$. The goal is to decide, for which $1 \leq i \leq N$, $v \in S_i$. Naturally, we assume $v \in \bigcup_{i=1}^N S_i$ $\forall v \in V$.
\fi

\iftr
\subsection{Transformers}
\fi

\textbf{Transformer}~\cite{vaswani2017attention} harnesses the multi-head attention mechanism in order to overcome the inherent sequential design of recurrent neural networks (RNNs)~\cite{rnn1, rnn2, rnn3}. Transformer has been effectively applied to different ML tasks, including but not limited to image recognition and time-series forecasting~\cite{vit, tft}. It has also laid the ground for many recent advances in generative AI~\cite{gpt, gpt3}.
We focus on the encoder-only architecture, following past DGRL works~\cite{dygformer}.
Let $x$ $=$ ($x_1$, \ldots $x_n$) be a sequence of $n$ $d$-dimensional inputs $x_i \in \mathbb{R}^d$. We consider the input matrix $X \in R^{n \times d}$, a single input token is represented as an individual row in $X$. \hl{Transformer details are in Appendix~A.}

\marginpar{\large\vspace{-1em}\colorbox{yellow}{\textbf{Audi}}\\\colorbox{yellow}{\textbf{(2.2)}}}

\if 0
\begin{wrapfigure}{r}{0.5\textwidth}
 \centering
 \vspace{-25pt}
 \includegraphics[width=0.5\textwidth]{transformer}
 \caption{Architecture of the Transformer model. (Figure taken from \cite{vaswani2017attention})}
 \label{fig:transformer}
 \vspace{-25pt}
\end{wrapfigure}
\fi

\iftr
\subsection{Block-Recurrent Transformer}
\label{sec:brt}
\fi

The runtime of   Transformer   increases quadratically with the length of the input sequence. As such, many efforts have been made to make it more efficient \cite{fastformer, flashatt, transxl, comptrans, swa}. Some of these efforts culminated in the \textbf{Block-Recurrent Transformer} \cite{brt}, a model which tries to bring together the advantages of Transformers and LSTMs \cite{lstm}, a special kind of RNN. As this is one of the principal building blocks of our model, we will briefly describe its architecture in the following.

The Block-Recurrent Transformer is essentially an extension of the Transformer-XL model \cite{transxl} and the sliding-window attention \cite{swa} mechanism. Long input sequences are divided into segments of size $S$, and further divided into blocks of size $B$. Following the sliding-window attention mechanism, instead of applying attention to the whole sequence, the Block-Recurrent Transformer applies it on each block individually. The elements of each block may attend to recurrently computed state vectors as well. With $B$ state vectors, we get attention matrices of size $B \times 2B$. As $B$ is constant, the cost of applying attention on each block is linear with respect to the segment size. This improves on the aforementioned quadratic runtime of the vanilla Transformer. \hl{More BRT details are in Appendix A.}

\marginpar{\large\vspace{-1em}\colorbox{yellow}{\textbf{Audi}}\\\colorbox{yellow}{\textbf{(2.2)}}}

\marginpar{\large\vspace{5em}\colorbox{yellow}{\textbf{All}}\\\colorbox{yellow}{\textbf{reviewers}}}

\marginpar{\large\vspace{5em}\colorbox{yellow}{\textbf{All}}\\\colorbox{yellow}{\textbf{reviewers}}}

\begin{figure}[hbtp]
\vspaceSQ{-1em}
 \centering
 \includegraphics[width=0.95\textwidth]{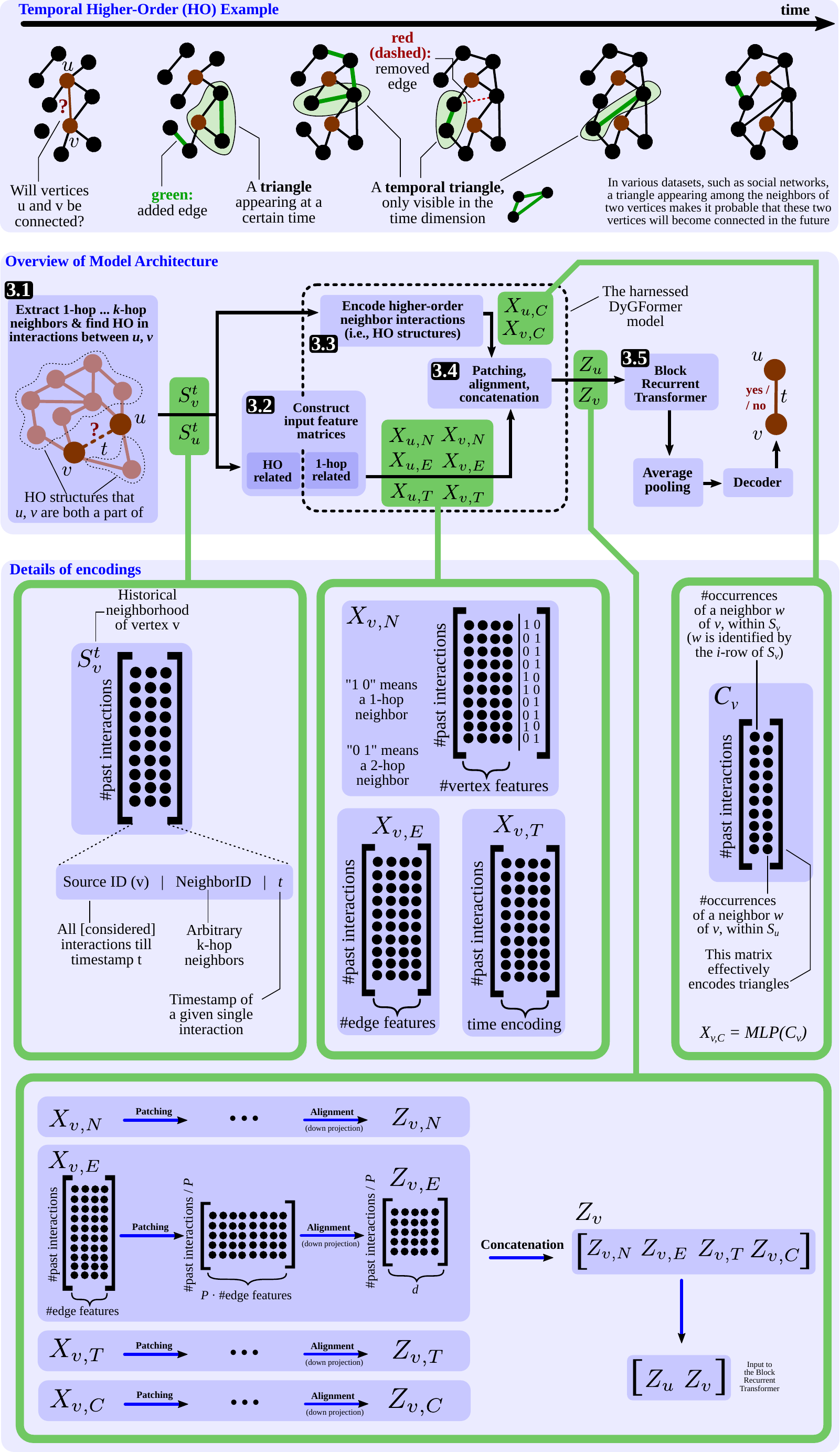}
\caption{\hl{Illustration of a temporal higher-order example and an overview of the HOT model.}}
\label{fig:hot}
\vspaceSQ{-1em}

\end{figure}

\section{The \nameAS Model}

We now present the design of \nameA, see Figure~\ref{fig:hot}. The key idea behind \nameAS is to harness the HO graph structures within the temporal dimension, and combine them with an efficient Transformer design applied to the temporal dimension modeled with tokens. For this, we extend the template design proposed by the DyGLib library~\cite{dygformer} that was provided as a setting where DyGFormer is implemented; it offers structured training and evaluation pipelines for dynamic link prediction.
We extend DyGLib and its DyGFormer design with the ability to harness HO and the hierarchical Transformers; we pick the Block-Recurrent Transformer (BRT) as a specific design of choice. 

Overall, \nameAS computes node representations within its encoder module; these representations are then leveraged to solve the downstream link prediction tasks using a suitable decoder.
Let $u$, $v$ be nodes in some CTDG and $t$ a timestamp. Given this, the model first extracts and appropriately encodes the higher-order neighbors of each vertex (Section~\ref{sec:model-ho-neighbors}), followed by constructing input feature matrices (Section~\ref{sec:model-input}). Then, the input matrices are augmented by encoding the selected HO interactions (Section~\ref{sec:model-ho-interactions}). After that, certain adjustments are made, such as passing the matrices via MLPs (Section~\ref{sec:model-patching}), followed by plugging in BRT (Section~\ref{sec:model-brt}).

\subsection{Extracting Higher-Order Neighbors}
\label{sec:model-ho-neighbors}

The model relies on the historical 1-hop and 2-hop interactions of the nodes $u$ and $v$ before $t$ to make its predictions.
The set of \textbf{1-hop} interactions of a vertex $u$ contains all tuples $(u, u', t')$ for which there is an interaction between nodes $u$ and $u'$ at timestamp $t'<t$. To enable a trade-off between memory consumption and accuracy, we only add the \emph{$s_1$ most recent} interactions from the set of 1-hop interactions to the list of considered interactions $S_u$ for $u$.
Then, we form the set of \textbf{2-hop} interactions. For each 1-hop interaction tuple $(u, u', t')$ in $S_u$, we consider the interactions $(u', u'', t'')$ with $t''<t$. We add $(u, u'', t'')$ for the $s_2$ most recent such interactions to the list of interactions $S_u$.
This scheme is further generalized to arbitrary $k$-hop neighbourhoods by iterating the construction processes. In our experiments, we focus on 1-hop and 2-hop neighborhoods.

\hl{The parameters $s_1$ and $s_2$ (and any other $s_k$ if one uses $k > 2$) introduce a tradeoff between accuracy vs.~preprocessing time and memory overhead. Specifically, larger values of any $s_i$ result in a more complete view of the temporal graph structure that is encoded into the HOT model. This usually increases the accuracy of predictions. On the other hand, they also increase the preprocessing as well as the memory overhead. This is because both the size $|S_u|$ and the preprocessing time are $O(s_1 s_2)$ (or $O(\prod_{i \in \{1, ..., k\}} s_i )$ for higher $k$).}

\subsection{Constructing Input Feature Matrices}
\label{sec:model-input}

In the next step, the model constructs neighbor and link encodings based on the raw node and link features of the CTDG. For node $u$, it constructs the matrices $X_{u, N} \in \mathbb{R}^{|S_u| \times d_N}$ and $X_{u, E} \in \mathbb{R}^{|S_u| \times d_E}$, where $d_N$ and $d_E$ are the dimensions of the node and edge feature vectors, respectively. 
The matrix $X_{u, N}$ is further extended to include a one-hot encoding to differentiate 1-hop neighbours from 2-hop neighbours. Specifically, two values $b_1$ and $b_2$ are appended to every row in $X_{u, N}$. We set $b_1 = 1$ and $b_2 = 0$ if the corresponding node is a 1-hop neighbour, and $b_1 = 0$ and $b_2 = 1$ if the corresponding node is a 2-hop neighbour; $d_N$ is updated accordingly.

\marginpar{\large\vspace{2em}\colorbox{yellow}{\textbf{1U56}}\\\colorbox{yellow}{\textbf{(2.1)}}}

For positional encodings, we rely on the scheme introduced in the TGAT model, reused in DyGLib. Here, for some timestamp $t'$, the time interval encoding for the time interval $\Delta t' = t - t'$ is given by 
%
%\vspace{-0.5em}
\begin{gather} 
\sqrt{{1}/{d_T}} \left[ \cos \left( w_1 \Delta t' \right), \sin \left( w_1 \Delta t' \right), \ldots, \cos \left( w_{d_T} \Delta t' \right), \sin \left( w_{d_T} \Delta t' \right) \right],
\end{gather}
%\vspace{-0.5em}
%
where $w_1$, \ldots, $w_{d_T}$ are trainable weights. The model computes this for every interaction in $S_u$, forming $X_{u, T}$. All these matrices are constructed for $v$ analogously.

\subsection{Encoding Higher-Order Neighbor Interactions} 
\label{sec:model-ho-interactions}

\marginpar{\large\vspace{2em}\colorbox{yellow}{\textbf{Audi}}\\\colorbox{yellow}{\textbf{(2.1)}}}

We next extend the DyGFormer's neighborhood encoding into the HO interactions. 
\hl{For this, we introduce matrices $C_u$, which -- for each vertex $u$ -- determine the count of neighbors shared by $u$ and any other vertex $v$ interacting with $u$, enabling us to encode temporal triangles containing $u$ and $v$ (in the following description, we focus on triangles for concreteness; other HO structures are enabled by considering neighbors beyond 1 hop).}
\hl{Specifically, for any vertex $u$, the $i$-th row of the matrix $C_u \in \mathbb{R}^{|S_u| \times 2}$ contains two numbers. The first one is the number of occurrences of a neighbor~$w$ of $u$ ($w$ is identified by the $i$-row of $S_u$) within $S_u$. The second number is the count of occurrences of $w$ within $S_v$, i.e., the temporal neighbourhood of $v$}. Then, we project these \hl{vectors of occurrences} onto a $d_C$-dimensional feature space using MLPs with one ReLU-activated hidden layer. The output are the two matrices $X_{u, C} \in \mathbb{R}^{|S_u| \times d_C}$, $X_{v, C} \in \mathbb{R}^{|S_v| \times d_C}$.
%
% it extracts matrices $C_u \in \mathbb{R}^{|S_u| \times 2}$ and $C_v \in \mathbb{R}^{|S_v| \times 2}$ by counting the occurrences of each neighbour in $S_u$ and $S_v$, respectively.
%
\if 0
We make this clear with an example from the DyGFormer paper. Given the historical neighbourhoods of $u$ and $v$, $[a$, $b$, $a]$ and $[b$, $b$, $a$, $c]$, the frequencies of $a$, $b$ and $c$ are $[2, 1]$, $[1, 2]$ and $[0, 1]$ respectively, where the first element refers to the frequency of the neighbour in the neighbourhood of $u$, and the second to the frequency of the neighbour in the neighbourhood of $v$. We then have, $C_u$ $=$ $[[2$, $1]$, $[1$, $2]$, $[2$, $1]]$ and $C_v$ $=$ $[[1$, $2]$, $[1$, $2]$, $[2$, $1]$, $[0$, $1]]$.
\fi
%
$X_{u, C}$ is then computed from $C_u$ as follows ($X_{v, C}$ is computed analogously): 
\begin{gather}
    X_{u, C} = \text{MLP}_0(C_u[:, 0]) + \text{MLP}_1(C_u[:, 1]) \in \mathbb{R}^{|S_u| \times d_C}.
\end{gather}
%
%, where $\text{MLP}(X) = \text{ReLu}(XW_1 + b_1)W_2 + b_2$; $W_1 \in \mathbb{R}^{1 \times d_C}$, $b_1 \in \mathbb{R}^{d_C}$, $W_2 \in \mathbb{R}^{d_C \times d_C}$, and $b_2 \in \mathbb{R}^{d_C}$ are trainable parameters.

\marginpar{\large\vspace{-2em}\colorbox{yellow}{\textbf{1U56}}\\\colorbox{yellow}{\textbf{(2.1)}}}

If $u$ and $v$ have a common 1-hop neighbour, then the subgraph induced by those three nodes is a triangle. Thus, this encodes the information about $u$ and $v$ being a part of this triangle into the harnessed feature matrix. More generally, by considering the common $k$-hop neighbors the model can encode any cycle containing nodes $u$ and $v$ of length up to $2k+1$. By extension, it can also encode any structure that is a conjunction of such cycles. This further increases the scope of the HO structures taken into account. 
%Considering higher-order neighbors would allow the model to encode longer cycles and, more generally, structures consisting of a conjunction of cycles. %In \nameA, by considering higher-order neighbourhoods, we encode the information of any cycles containing the nodes $u$ and $v$. By extension, we also encode any structure, which is a conjunction of such cycles. This further increases the scope of the HO structures taken into account. 

%Let us consider the matrix $C_v$ from the example above, and analyse its first column $[1, 1, 2, 0]^T$. Recall, that a non-zero element here denotes the presence of the respective neighbour of $v$ in the neighbourhood of $u$. As such, it also represents the presence of cycles containing that very node. Note, that the value exactly corresponds to the number of such cycles containing the respective interaction. Hence, in this case, there is only one cycle containing the more recent interaction $(v, b, t_b')$, one cycle containing the older interaction $(v, b, t_b'')$, $2$ cycles containing the interaction $(v, a, t_a)$ and no cycles containing the interaction $(v, c, t_c)$. While we do not encode the length of these cycles, we can vaguely infer it from the position of the element that represents it. Elements in the first rows will correspond to 1-hop neighbours, and, as such, encode shorter cycles (triangles), while elements in the last rows will likely represent longer cycles. 

\subsection{Patching, Alignment, Concatenation} 
\label{sec:model-patching}

Before feeding $X_{\cdot, \cdot}$ into the selected efficient Transformer, we harness several optional transformations from DyGLib, which further improve the model performance and memory utilization.
First, the patching technique bundles multiple rows of a matrix together into one row, so as to reduce the size of the input sequence. 
Then, alignment reduces the feature dimension of this input by projecting each row onto a smaller dimension. Let us examine the scheme on the matrix $X_{u, N}$. The model constructs $M_{u, N} \in \mathbb{R}^{l_u \times P \cdot d_N}$ by dividing $X_{u, N}$ into $l_u = \left\lceil \frac{|S_u|}{P} \right\rceil$ patches and flattening $P$ temporally adjacent encodings. The same procedure is applied to the other matrices $X_{\cdot, \cdot}$. Intuitively, this patching scheme can be seen as making the timestamps ``coarses'', e.g., if the patching factor is 4, a single row in the resulting patched matrix contains the information about previous 4 timestamps.

The encodings then need to be aligned to a common dimension $d$. For the matrix $M_{u, N}$, we have 
\begin{gather}
Z_{u, N} = M_{u, N} W_N + b_N \in \mathbb{R}^{l_u \times d},
\end{gather}
where $W_N \in \mathbb{R}^{P \cdot d_N \times d}$ and $b_N \in \mathbb{R}^{d}$ are trainable parameters. The matrices $Z_{u, E}$, $Z_{u, T}$, and $Z_{u, C}$ are extracted identically from the matrices $M_{u, E}$, $M_{u, T}$, and $M_{u, C}$. The same applies to the matrices belonging to node $v$. 

Finally, the extracted matrices are concatenated horizontally into

\marginpar{\large\vspace{-5em}\colorbox{yellow}{\textbf{1U56}}\\\colorbox{yellow}{\textbf{(2.1)}}}

\vspace{-0.5em}
\begin{gather}
Z_u = Z_{u, N} \| Z_{u, E} \| Z_{u, T} \| Z_{u, C} \in \mathbb{R}^{l_u \times 4d},\quad Z_v = Z_{v, N} \| Z_{v, E} \| Z_{v, T} \| Z_{v, C} \in \mathbb{R}^{l_v \times 4d}.    
\end{gather}

Finally, the two matrices $Z_u$ and $Z_v$ must also be concatenated, before being fed to the next stage of the pipeline. In \nameA, we concatenate the matrices horizontally, as opposed to the vertical concatenation in DyGFormer. This may prevent the model from falsely interpreting all events encoded in one of the matrices as happening before or after all events encoded in the other matrix.

\iftr
While the DyGFormer would concatenate the two matrices vertically, i.e., the sequence $Z_u \in \mathbb{R}^{l_u \times 4d}$ would be extended with $Z_v \in \mathbb{R}^{l_v \times 4d}$ to make $Z \in \mathbb{R}^{(l_u + l_v) \times 4d}$, the HOT concatenates the matrices horizontally. The resulting matrix belongs to $\mathbb{R}^{\max \{l_u, l_v\} \times 8d}$. Note, that the dimensions $l_u$ and $l_v$ may differ. In this case, the matrix with the lowest number of rows is padded with zeros to match the size of the larger matrix. The concatenated matrix is fed directly to BRT.
\fi

\subsection{Harnessing Temporal Hierarchy with Block-Recurrent Transformer} 
\label{sec:model-brt}

Finally, we harness a hierarchical Transformer to minimize memory required for keeping the HO temporal structures.
 The BRT divides $Z$ into $B$ blocks and applies attention locally on each individual block. Additionally, each block is cross-attended with a recurrent state, which allows the element in one block to attend to a summary of the elements in the previous blocks. The outputs of each block are then concatenated into a matrix $H$, which shares its dimensions with the input $Z$. Finally, temporal node representations for nodes $u$ and $v$ are computed with the average pooling layer.

\subsection{Computational Cost}
\label{sec:model-analysis}

\iftr
We bound the overall cost to construct the matrices that are fed to the Transformer.
\fi
Let $s=\prod_i s_i$, such that $|S_u|\in O(s)$ and $|S_v|\in O(s)$ and let $\Delta$ be the largest number of interactions among vertices $u$ and $v$. Then, the cost to compute $S_u$ and $S_v$ is $O(s\Delta \log \Delta)$.
% Assuming constant-time access to the most recent events for each vertex, the cost to construct $S_u$ and $S_v$ is $O(s)$. Such a constant-time access can be maintained by keeping a ring-buffer of events for each neighbor as they arrive. 
%
The cost to construct the $Z_{u}$ and $Z_v$ is dominated by the alignment process, which costs $O(s d_N d)$ for the node matrix $Z_{u, N}$. The cost is analogous for the other matrices constituting $Z_u$ and $Z_v$.
The remaining costs are that of the BRT model for feature dimension 8d, which is, in particular, linear in the number of interactions $s$. The attention mechanism in a block-recurrent Transformer costs $O(sdB/P)$~\cite{brt}, compared to $\Theta(s^2d/P)$ in the vanilla Transformer. As the block size $B$ approaches the number of interactions $s$, the cost of the BRT attention approaches that of the vanilla Transformer.

%\vspaceSQ{-1em}
\section{Evaluation}
\label{sec:eval}

We now illustrate the advantages of \nameAS over the state of the art.
%
% Our analysis comes with a large evaluation space. Thus, we show selected representative results; full data is in the appendix due to space constraints.

%\vspaceSQ{-0.5em}
\subsection{Experimental Setup}
%\vspaceSQ{-0.25em}

We follow recent established methodologies for evaluating DGRL~\cite{evaluation}. We list the most relevant information here, and include details of model parameters and related information in the appendix.
We use the standard evaluation metrics, namely the \textbf{Average Precision (AP)} (using the scikit implementation~\cite{scikit}) and the \textbf{Area Under the ROC (AUC)}.

\iftr
\subsubsection{Comparison Baselines}
\fi

We use all major state-of-the-art baselines for DGRL on CTDGs; these are TGN~\cite{tgn}, CAWN~\cite{cawn}, TCL~\cite{tcl}, GraphMixer~\cite{graphmixer}, DyGFormer~\cite{dygformer}), as well as a purely memorisation-based approach (EdgeBank~\cite{evaluation}). Note that our analyses confirm recent findings that illustrate the superiority of DyGFormer among these baselines~\cite{dygformer}. 
%
%Hence, to make the plots more clear, we only compare to DyGFormer, and move the outcomes of other baselines to the appendix.

\iftr
\subsubsection{Evaluation Settings \& Negative Edge Sampling}
\fi 

We consider both \textbf{transductive} and \textbf{inductive} setting. In the former, the whole graph structure is visible during training. In the latter, we test the dynamic link prediction and the dynamic node classification on the graph structure that was not visible during training.

\marginpar{\large\vspace{1em}\colorbox{yellow}{\textbf{gy2x}}\\\colorbox{yellow}{\textbf{(2)}}}

\iftr
For evaluation, one requires negative edges, i.e., the edges that determine the absent links, which the model should detect as non-existing. 
\fi
\hl{Recent work on benchmarking dynamic GRL~\mbox{\cite{dygformer}} illustrates the importance of evaluating different sampling schemes beyond plain random sampling. We follow this approach and use all three discussed sampling strategies, showing that HOT achieves competitive results over the state of the art \emph{regardless of how links are sampled}.}
First, we use \textbf{random sampling} (RNES), which is widely used in most dynamic link prediction evaluations. In RNES, when observing some positive edge sample, we generate a negative one by changing its destination node to some random vertex.
Second, we use a recent scheme called \textbf{historical sampling} (HNES)~\cite{evaluation}. In many datasets, it is common to observe repeated interactions (i.e., edges that appear and disappear several times). With random sampling, most negative edge queries will be new to the model (i.e., unobsered in the past). As such, it becomes easy to discard these edges, instead of discarding the repeated edges. In historical sampling, we sample negative edges from the set of previously observed edges, which do \emph{not} appear in the current timestamp (if we cannot sample enough edges this way, the rest is sampled using random sampling).
Third, we also use \textbf{inductive sampling} (INES)~\cite{evaluation}, in which we sample edges from those that were not seen during training and do not appear in the current timestamp, i.e., new edges. As before, if we cannot sample enough edges this way, the rest is sampled using the random sampling technique. This scheme produces a metric to better evaluate the model's ability to induce new relations.

All three above sampling schemes can be considered for both transductive or inductive evaluation setting.
Note that we do not consider the historical and the inductive sampling techniques in the inductive evaluation setting. This is because the sampled set becomes quite small and unlikely to produce meaningful results. Additionally, consider that both techniques are equivalent in the inductive setting. This is clearly visible in the results obtained by DyGFormer~\cite{dygformer}.

\if 0

% Maciej: not good enough, pity!

\vspaceSQ{-0.5em}
\subsection{Enhancing Node Classification with Higher Order}
\vspaceSQ{-0.25em}

\afonso{Note: direct extraction from thesis paper. May need some adaptation.}

As only the Wikipedia and Reddit datasets contain categorical labels, we can only test node classification on these datasets. Furthermore, while we could present both AP and AUC-ROC scores, we choose to rely on the latter. The reasoning behind this is the rarity of positive labels in both datasets. This leads to generally low but, most importantly, very sensitive AP scores. 

The results are plotted in Figure \ref{fig:nc}. On the Wikipedia dataset the model achieves comparable results to most other baselines, lagging only slightly behind the DyGFormer. On the reddit dataset, however, the moddel is able to beat the DyGFormer and lie just slightly behind the SOTA set by TGAT.

\begin{figure}[ht]
 \centering
 \includegraphics[width=0.7\textwidth]{NC}
\caption{AUC (\%) scores for node classification. Baseline results are taken from \cite{dygformer}. The HOT's Wikipedia results are averaged over 5 runs using an early stopping patience of 2, while the Reddit results are averaged over 3 runs using an early stopping patience of 5. The configurations chosen for the model are the one that achieved the best results in the dynamic link prediction task.}
\label{fig:nc}
\end{figure}

\fi

%\vspaceSQ{-0.5em}
\subsection{Analysis of Performance}
\label{sec:lp}
\vspaceSQ{-0.25em}

\begin{figure}[hbtp]
 \centering
 \includegraphics[width=1.0\textwidth]{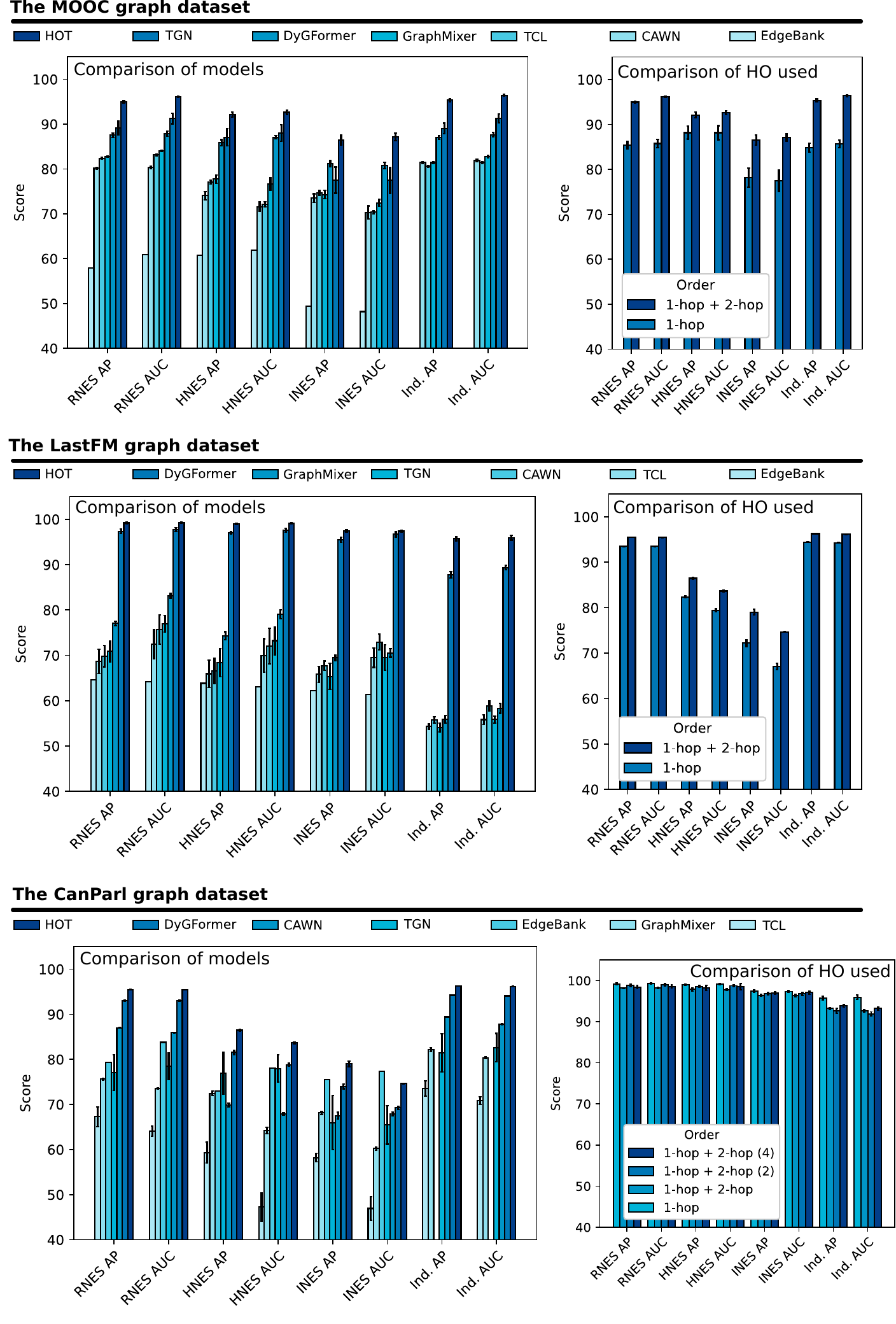}
\caption{AP (\%) and AUC (\%) scores on the MOOC, LastFM and CanParl datasets using the various negative edge sampling techniques (RNES, HNES, INES) in the transductive setting, and using the random negative edge sampling technique in the inductive setting (Ind). Baseline results are the best ones provided by~\cite{dygformer}.}
\label{fig:main-data}
\end{figure}

We illustrate the comparison of the performance of different models in Figure~\ref{fig:main-data}.
In the MOOC graph dataset, the model successfully leverages 2-hop interactions to make its predictions more accurate. This leads to the best results over all the evaluation metrics and negative edge sampling strategies. These advantages hold for both inductive and inductive settings.
Similar performance patterns can be seen for the LastFM dataset, where higher-order neighbors also enable obtaining more powerful predictions.
As for the Can.~Parl.~dataset, the results obtained without the higher-order structure are already quite high, and the additional graph structural information does not seem to be adding significant amount of value to the model in this case. Furthermore, it is possible that the decay in performance with higher $s_2$ values originates from the added noise that comes with considering more information. In any case, \nameAS still ensures the highest scores for both the AP and the AUC metric.

Another factor that benefits \nameA, particularly in comparison with the DyGFormer, is the horizontal concatenation of the matrices $Z_u$ and $Z_v$, as opposed to the vertical concatenation in the DyGFormer case. This horizontal concatenation strategy forces the attention modules to consider the elements in the same position (in both matrices/sequences) together, as one element. We conjecture that this prevents the model from a potential false interpretation, in which all events encoded in one of the $Z_u$ and $Z_v$ matrices take place before all events encoded in the other matrix. Thus, we conjecture that this additionally allows the model to better infer the chronological position elements in the sequence. 

\if 0
While the model exhibits good results in the datasets above and some others, such as UCI, and US Legis., this is not always the case. In some datasets the model is comparable to previous approaches and in others it falls short. It is a common occurrence that the model cannot leverage any higher order information and defaults to results that resemble the DyGFormer. We can only deduce the cause of this phenomenon. Either these datasets don't exhibit any higher order patterns 
\fi

\iftr
\textbf{Summary}
Overall, \nameAS offers advantages in accuracy over all the considered baselines, for all the considered metrics, in different datasets.
\fi

\marginpar{\large\vspace{1em}\colorbox{yellow}{\textbf{Audi}}\\\colorbox{yellow}{\textbf{(1)}}\\\colorbox{yellow}{\textbf{1U56}}\\\colorbox{yellow}{\textbf{(1)}}}

%\vspaceSQ{-0.5em}
\subsection{{Analysis of Higher-Order (HO) Characteristics}}
\label{sec:lp}
\vspaceSQ{-0.25em}

\hl{The benefits of harnessing HO structures are clearly visible in the rightmost plots of Figure~\mbox{\ref{fig:main-data}}. There, we fix $s_1$ and vary $s_2$. For the MOOC and LastFM, we consider $s_2 \in \{0, 1\}$; for CanParl, we consider $s_2 \in \{0, 1, 2, 4\}$. By setting $s_2 = 0$, we exclude any HO  structures beyond triangles from the model's consideration (the results in this setting correspond to the bars labeled "1-hop"). Overall, CanParl does not benefit from higher values of $s_2$. Contrarily, for the other two datasets, the model clearly benefits from the HO structures. Setting $s_2 = 1$, i.e., considering not only triangles, but any cycles of up to 5 nodes, further improves performance.}

%\vspaceSQ{-0.5em}
\subsection{Analysis of Memory Consumption}
\label{sec:lp}
\vspaceSQ{-0.25em}

\begin{figure}[t]
 \centering
 \includegraphics[width=1.0\textwidth]{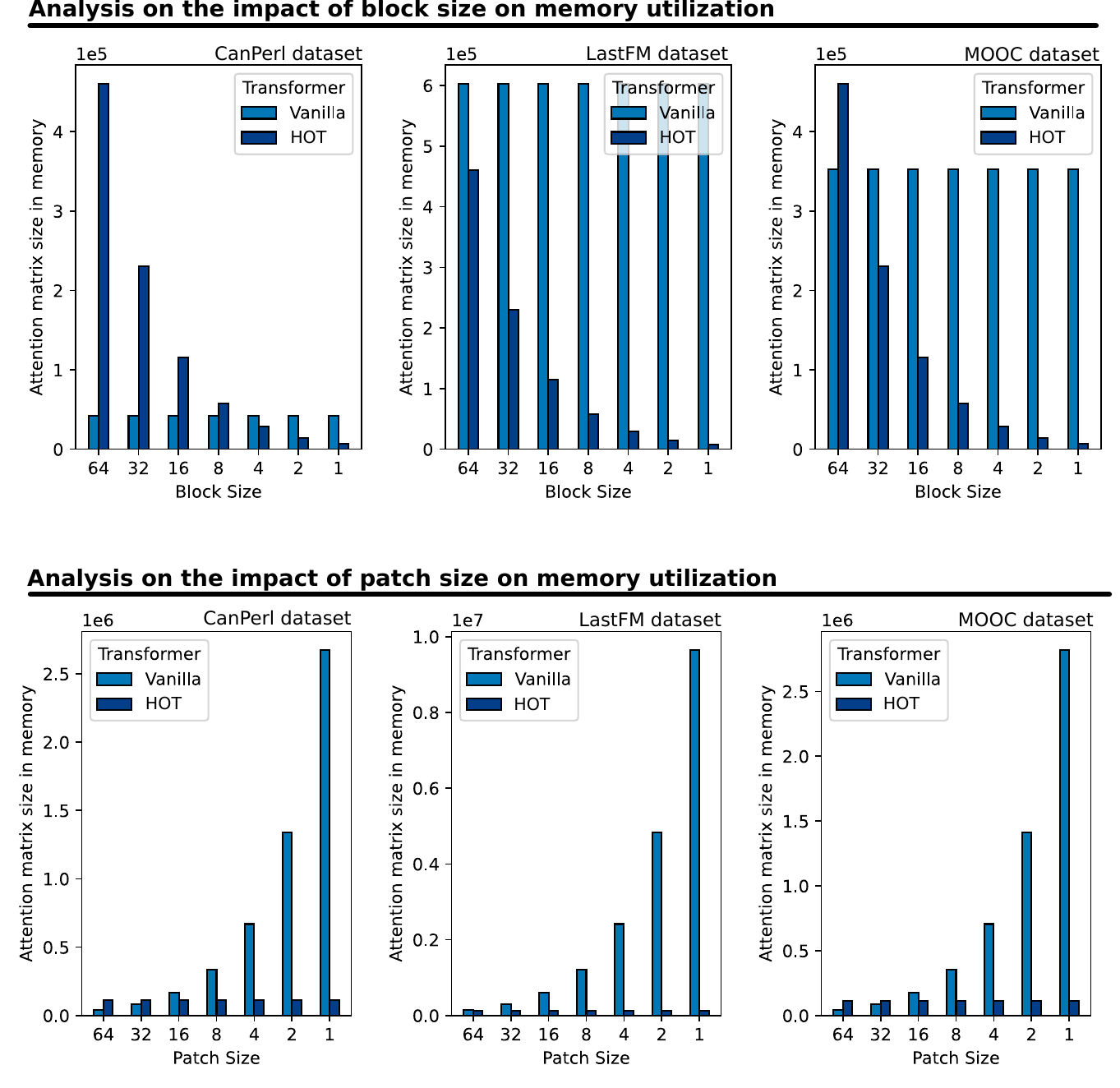}
\caption{The analysis of the impact of the block and patch size on memory utilization.}
\label{fig:mem-analysis}
\end{figure}

\marginpar{\large\vspace{1em}\colorbox{yellow}{\textbf{Audi}}\\\colorbox{yellow}{\textbf{(3)}}}

\hl{We also investigate in more detail the impact of the block size $B$ (see Section~\mbox{\ref{sec:model-brt}}) and the patch size $P$ (see Section~\mbox{\ref{sec:model-patching}}) on the memory consumption. The results are in Figure~\mbox{\ref{fig:mem-analysis}} and they show the size needed for the attention matrix in Vanilla Transformer, compared to \nameA.
First, assuming that blocks are processed sequentially as in RNNs, reducing $B$ also decreases the needed memory (for a fixed $P$). While large values of $B$ inflate the total memory beyond what is needed for the Vanilla Transformer, smaller blocks result in much less memory needed. This comes with a tradeoff, and the latency to process the model increases linearly with the number of blocks. This however can be alleviated with schemes designed to specifically speed up the RNN processing.
The patch size $P$ has a similar effect: the smaller it is, the more memory is required for Vanilla Transformer (for a fixed $B$), due to the patching flattening (Section~\mbox{\ref{sec:model-patching}}). Hence, selecting $B$ and $P$ is a design choice that would impact both the needed memory and performance. In our implementation, we offer a set of scripts that offer a quick assessment of the required memory, facilitating the design of schemes based on \nameA.
Overall, based on the given input sequence length $S$, the patch size $P$, and the block size $B$, the total amount of elements in attention matrices of the vanilla Transformer can be assessed as $(3S / P) \cdot (8d)$. Furthermore, considering just one block, the attention matrix based on BRT needs about $(9 \cdot 2B) \cdot (8d)$ elements.}
%
% We collected the average sequence lengths $S$ of each dataset (with $s_2 = 20$). The figures below plot these estimates with varying values of $P$ (with $B = 16$) and $B$ (with $P$ the patch size we chose for the respective dataset).

%\afonso{Note: This is all assuming we can keep ONLY ONE block in memory, do the computations, store the output, and then continue with the next block.}

\if 0
\begin{figure}[ht]
 \centering
 \includegraphics[width=0.3\textwidth]{CanParlMemBlock}
\caption{}
\end{figure}

\begin{figure}[ht]
 \centering
 \includegraphics[width=0.3\textwidth]{lastfmMemBlock.pdf}
\caption{}
\end{figure}

\begin{figure}[ht]
 \centering
 \includegraphics[width=0.3\textwidth]{moocMemBlock}
\caption{}
\end{figure}

\begin{figure}[ht]
 \centering
 \includegraphics[width=0.3\textwidth]{CanParlMemPatch}
\caption{}
\end{figure}

\begin{figure}[ht]
 \centering
 \includegraphics[width=0.3\textwidth]{lastfmMemPatch}
\caption{}
\end{figure}

\begin{figure}[ht]
 \centering
 \includegraphics[width=0.3\textwidth]{moocMemPatch}
\caption{}
\end{figure}
\fi

\enlargeSQ
\enlargeSQ

\section{{Discussion}}

\marginpar{\large\vspace{3em}\colorbox{yellow}{\textbf{ZJkb}}\\\colorbox{yellow}{\textbf{(c)}}}

%\vspaceSQ{-0.5em}
\subsection{{Applications of HOT Beyond Link Prediction}}
\label{sec:limit}
%\vspaceSQ{-0.25em}

\hl{The generic structure of HOT ensures that it can be straightforwardly extended towards other dynamic graph ML tasks, including edge, node, or graph classification or regression.}

\hl{To illustrate this on a concrete example, we implemented \emph{dynamic node classification} within HOT. Here, we are given a CTDG ($G^{(0)}$, $T$), a timestamp $t \in \mathbb{N}$, a node $v \in V$, and a set of $N \in \mathbb{N}$ classes $\{S_i\}_{i=1}^N$. The goal is to decide, for which $1 \leq i \leq N$, $v \in S_i$. Naturally, we assume $v \in \bigcup_{i=1}^N S_i$ $\forall v \in V$. To solve this task in HOT, we employ transfer learning, i.e., we harness the parameters from the dynamic link prediction task to compute suitable dynamic node representations, and then train a suitable MLP decoder on those representations. The preliminary evaluation indicate performance comparable to, or better than DyGFormer on -- respectively --  the Wikipedia and the Reddit datasets. A more detailed investigation into the model design for dynamic node classification, as well as more extensive evaluations, are future work. However, our preliminary results indicate potential in harnessing the HO structures in the general dynamic GRL setting.}

\hl{Other dynamic GRL tasks could be achieved using established GRL methods. For example, graph classification could be used by applying pooling on top of the edge and node classification~\mbox{\cite{kazemi2020representation, barros2021survey, skarding2021foundations, sizemore2018dynamic, xue2022dynamic, han2021dynamic}}.}

\marginpar{\large\vspace{3em}\colorbox{yellow}{\textbf{ZJkb}}\\\colorbox{yellow}{\textbf{(b)}}}

%\vspaceSQ{-0.5em}
\subsection{{Limitations of HOT}}
\label{sec:limit}
%\vspaceSQ{-0.25em}

\hl{The limitations of HOT are largely analogous to those of any HO scheme. First, the time complexity required to search for HO structures may limit the size of the graphs or the length of the temporal history to be considered. Here, HOT's harnessing of efficient Transformers alleviates the time and storage complexity needed for the on-the-fly predictions. Still, HOT needs to search for HO structures to construct the input feature matrices; however, it is a one-time preprocessing overhead for a given dataset.
Second, utilizing ``too much'' of HO may introduce excessive amounts of noise and ultimately lower the accuracy. This limitation is also generic to HO learning, it is commonly alleviated by appropriately limiting the size of the harnessed HO structures~\mbox{\cite{zhang2018link}.}}
\section{Related Work}
  
Our work touches on many areas. We now briefly discuss related works.

\textbf{Graph Neural Networks and Graph Representation Learning}
Graph neural networks (GNNs) emerged as a highly successful part of the graph representation learning (GRL) field~\cite{hamilton2017representation}. Numerous GNN models have been developed~\cite{wu2020comprehensive, zhou2020graph, zhang2020deep, chami2020machine, hamilton2017representation, bronstein2017geometric, besta2021motif, gianinazzi2021learning, scarselli2008graph, besta2022parallel}, including convolutional~\cite{kipf2016semi, hamilton2017inductive, wu2019simplifying, xu2018powerful, sukhbaatar2016learning}, attentional~\cite{monti2017geometric, velivckovic2017graph, thekumparampil2018attention, besta2023high}, message-passing~\cite{wang2019dynamic, bresson2017residual, sanchez2020learning, battaglia2018relational, gilmer2017neural}, or -- more recently -- higher-order (HO) ones~\cite{abu2019mixhop, rossiHigherorderNetworkRepresentation2018, rossiHONEHigherOrderNetwork2018, benson2018simplicial, abu2019mixhop, morris2019weisfeiler, bodnar2021weisfeiler}.
%
% cook2006mining, jiang2013survey, horvath2004cyclic, chakrabarti2006graph
%
Moreover, a large number of software frameworks~\cite{wang2019deep, fey2019fast, li2020pytorch, zhang2020agl, jia2020improving, zhu2019aligraph, wang2020gnnadvisor, hu2020featgraph, wu2021seastar, ma2019neugraph, tripathy2020reducing, wan2022pipegcn, wan2022bns, zheng2021distributed, waleffe2022marius++}, and even hardware accelerators~\cite{kiningham2020greta, liang2020engn, yan2020hygcn, geng2020awb, kiningham2020grip} for processing GNNs have been introduced over the last years.
All these schemes target static graphs, with no updates, while in this work, we target the dynamic GRL, where graphs evolve over time. However, our model can be seamlessly used for a temporal static setting, where the graph comes with the available history of past updates.

\textbf{Dynamic Graph Representation Learning}
There have been many approaches to solving dynamic GRL in the past years.
JODIE~\cite{jodie} is an RNN-based approach for bipartite graphs. The model constructs and maintains embeddings of source and target nodes and updates these recurrently using different RNNs.
DyRep~\cite{dyrep} is an RNN-based approach, which keeps node representations and updates them using a self-attention mechanism, which dynamically weights the importance of different structures in the graph.
TGAT~\cite{tgat} computes node representations based on the static GAT model~\cite{velivckovic2017graph}, i.e., by aggregating messages from the temporal neighborhood of each node using self-attention. The main difference between TGAT and GAT is that timestamps are taken as input as well and taken into consideration after a suitable time encoding is found.
TGN~\cite{tgn} divides the model architecture into the message function, the message aggregator, the memory updater, and the embedding module. Given two nodes and a timestamp, the model fetches node representations from memory, aggregates them, and updates its memory. It then builds node embeddings for downstream tasks using a TGAT-like layer. 
CAWN~\cite{cawn} builds on the concept of anonymous walks~\cite{aw} and extends them to consider the time dimension. Given two nodes and a timestamp, the model collects causal anonymous walks starting from both nodes. They are then encoded using RNNs and finally aggregated for predictions.
TCL~\cite{tcl} first orders previous interactions of each node in some order that reflects both temporal and structural positioning (for any two pairs of nodes). The model then runs a Transformer on each of the nodes' previous interactions. These Transformers are coupled through cross-attention.
GraphMixer~\cite{graphmixer} is a simple architecture consisting of three modules: the link encoder, the node encoder, and the link classifier. The link encoder summarizes temporal link information using an MLP-mixer \cite{mlp-mixer}, and the node encoder captures node information using neighbour mean-pooling. The link classifier applies a 2-layer MLP on the output of the two other modules to formulate a prediction.
Finally, DyGFormer~\cite{dygformer} is an approach based on the Transformer model.
The model proposed in this work, \nameA, extends DyGFormer and outperforms all other baselines for different datasets thanks to harnessing the HO graph structures.

\iftr
\textbf{Transformers and Attention}
Transformers~\cite{vaswani2017attention} have had a huge impact on Natural Language Processing (NLP) and Computer Vision (CV) \cite{vit, tft, gpt, gpt3}. More specifically, they have performed exceptionally well on data of sequential nature, such as words in a sentence.
More recently, \emph{hierarchical Transformer models} were proposed, for example RWKV~\cite{peng2023rwkv}, Swin Transformer~\cite{liu2021swin}, hierarchical BERT~\cite{pappagari2019hierarchical}, Nested Hierarchical Transformer~\cite{zhang2022nested}, Hift~\cite{cao2021hift}, or Block-Recurrent Transformer~\cite{brt}.
Our model harnesses these developments into the domain of dynamic GRL.
\fi

\textbf{Dynamic and Streaming Graph Computing}
There also exist systems for processing dynamic and streaming graphs~\cite{besta2019practice, sakr2020future, choudhury2017nous, besta2021enabling} beyond GRL. \emph{Graph streaming frameworks} such as STINGER~\cite{feng2015distinger} or Aspen~\cite{dhulipala2019low} emerged to enable processing and analyzing dynamically evolving graphs. \emph{Graph databases} are systems used to manage, process, analyze, and store vast amounts of rich and complex graph datasets. Graph databases have a long history of development and focus in both academia and in the industry, and there has been significant work on them~\cite{besta2019demystifying, besta2023gdi, besta2023thegdi, angles2018introduction, davoudian2018survey, han2011survey, gajendran2012survey, gdb_survey_paper_Kaliyar, kumar2015domain, gdb_survey_paper_Angles}.  Such systems often execute graph analytics algorithms (e.g., PageRank) {concurrently} with graph updates (e.g., edge insertions). Thus, these frameworks must tackle unique challenges, for example effective modeling and storage of dynamic datasets, efficient ingestion of a stream of graph updates concurrently with graph queries, or support for effective programming model. Here, recently introduced \emph{Neural graph databases} focus on integrating Graph Databases with GRL capabilities~\cite{besta2022neural, ren2023neural}.
Our model could be used to extend these systems with dynamic GRL workloads.

\section{Conclusion}

Dynamic graph representation learning (DGRL), where graph datasets may ingest millions of edge updates per second, is an area of growing importance.
In this work, we enhance one of the most recent and powerful works in DGRL, the Transformer-based DGRL, by harnessing higher-order (HO) graph structures: k-hop neighbors and more general subgraphs. As this approach enables harnessing more information from the temporal dimension, it results in the higher accuracy of prediction outcomes. Simultaneously, it comes at the expense of increased memory pressure through the larger underlying attention matrix. For this, we employ a recent class of Transformer models that impose hierarchy on the attention matrix, picking Block-Recurrent Transformer for concreteness. This reduces memory footprint while ensuring -- for example -- 9\%, 7\%, and 15\% higher accuracy in dynamic link prediction than -- respectively -- DyGFormer, TGN, and GraphMixer, for the MOOC dataset. 

Our design illustrates that a careful combination of models and paradigms used in different settings, such as the Higher-Order graph structures and the Block-Recurrent Transformer, results in advantages in both performance and memory footprint in DGRL. This approach could be extended by considering other state-of-the-art Transformer schemes or HO GNN models.

\vspace{2em}

%\begin{ack}

\ifnbld

\section*{Acknowledgements}

We thank Hussein Harake, Colin McMurtrie, Mark Klein, Angelo Mangili, and the whole CSCS team granting access to the Ault and Daint machines, and for their excellent technical support. 
We thank Timo Schneider for help with infrastructure at SPCL.
This project received funding from the European Research Council (Project PSAP, No.~101002047), and the European High-Performance Computing Joint Undertaking (JU) under grant agreement No.~955513 (MAELSTROM). This project was supported by the ETH Future Computing Laboratory (EFCL), financed by a donation from Huawei Technologies. This project received funding from the European Union’s HE research and innovation programme under the grant agreement No. 101070141 (Project GLACIATION).

\fi

%\end{ack}

{
  \footnotesize
\bibliographystyle{ACM-Reference-Format}
\bibliography{references, references-afonso}


\newcommand{\SortNoop}[1]{}
\begin{thebibliography}{143}

%%% the Web address.

\ifx \showCODEN    \undefined \def \showCODEN     #1{\unskip}     \fi
\ifx \showDOI      \undefined \def \showDOI       #1{#1}\fi
\ifx \showISBNx    \undefined \def \showISBNx     #1{\unskip}     \fi
\ifx \showISBNxiii \undefined \def \showISBNxiii  #1{\unskip}     \fi
\ifx \showISSN     \undefined \def \showISSN      #1{\unskip}     \fi
\ifx \showLCCN     \undefined \def \showLCCN      #1{\unskip}     \fi
\ifx \shownote     \undefined \def \shownote      #1{#1}          \fi
\ifx \showarticletitle \undefined \def \showarticletitle #1{#1}   \fi
\ifx \showURL      \undefined \def \showURL       {\relax}        \fi
% invisible to TeX
\providecommand\bibfield[2]{#2}
\providecommand\bibinfo[2]{#2}
\providecommand\natexlab[1]{#1}
\providecommand\showeprint[2][]{arXiv:#2}

\bibitem[{Abu-El-Haija} et~al\mbox{.}(2019)]%
        {abu2019mixhop}
\bibfield{author}{\bibinfo{person}{Sami {Abu-El-Haija}}, \bibinfo{person}{Bryan
  Perozzi}, \bibinfo{person}{Amol Kapoor}, \bibinfo{person}{Nazanin
  Alipourfard}, \bibinfo{person}{Kristina Lerman}, \bibinfo{person}{Hrayr
  Harutyunyan}, \bibinfo{person}{Greg Ver~Steeg}, {and} \bibinfo{person}{Aram
  Galstyan}.} \bibinfo{year}{2019}\natexlab{}.
\newblock \showarticletitle{Mixhop: Higher-order graph convolutional
  architectures via sparsified neighborhood mixing}. In
  \bibinfo{booktitle}{\emph{international conference on machine learning}}.
  PMLR, \bibinfo{pages}{21--29}.
\newblock


\bibitem[Alvarez-Rodriguez et~al\mbox{.}(2021)]%
        {socNet3}
\bibfield{author}{\bibinfo{person}{Unai Alvarez-Rodriguez},
  \bibinfo{person}{Federico Battiston}, \bibinfo{person}{Guilherme~Ferraz de
  Arruda}, \bibinfo{person}{Yamir Moreno}, \bibinfo{person}{Matja{\v{z}} Perc},
  {and} \bibinfo{person}{Vito Latora}.} \bibinfo{year}{2021}\natexlab{}.
\newblock \showarticletitle{Evolutionary dynamics of higher-order interactions
  in social networks}.
\newblock \bibinfo{journal}{\emph{Nature Human Behaviour}} \bibinfo{volume}{5},
  \bibinfo{number}{5} (\bibinfo{year}{2021}), \bibinfo{pages}{586--595}.
\newblock


\bibitem[Ammar(2016)]%
        {ammar2016techniques}
\bibfield{author}{\bibinfo{person}{Khaled Ammar}.}
  \bibinfo{year}{2016}\natexlab{}.
\newblock \showarticletitle{Techniques and Systems for Large Dynamic Graphs}.
  In \bibinfo{booktitle}{\emph{SIGMOD'16 PhD Symposium}}. ACM,
  \bibinfo{pages}{7--11}.
\newblock


\bibitem[Angles and Gutierrez(2008)]%
        {gdb_survey_paper_Angles}
\bibfield{author}{\bibinfo{person}{Renzo Angles} {and} \bibinfo{person}{Claudio
  Gutierrez}.} \bibinfo{year}{2008}\natexlab{}.
\newblock \showarticletitle{{Survey of Graph Database Models}}.
\newblock \bibinfo{journal}{\emph{in ACM Comput. Surv.}} \bibinfo{volume}{40},
  \bibinfo{number}{1}, Article \bibinfo{articleno}{1} (\bibinfo{year}{2008}),
  \bibinfo{numpages}{39}~pages.
\newblock
\showISSN{0360-0300}
\urldef\tempurl%
\url{https://doi.org/10.1145/1322432.1322433}
\showDOI{\tempurl}


\bibitem[Angles and Gutierrez(2018)]%
        {angles2018introduction}
\bibfield{author}{\bibinfo{person}{Renzo Angles} {and} \bibinfo{person}{Claudio
  Gutierrez}.} \bibinfo{year}{2018}\natexlab{}.
\newblock \showarticletitle{{An Introduction to Graph Data Management}}.
\newblock In \bibinfo{booktitle}{\emph{Graph Data Management, Fundamental
  Issues and Recent Developments}}. \bibinfo{pages}{1--32}.
\newblock


\bibitem[Ba et~al\mbox{.}(2016)]%
        {ln}
\bibfield{author}{\bibinfo{person}{Jimmy~Lei Ba}, \bibinfo{person}{Jamie~Ryan
  Kiros}, {and} \bibinfo{person}{Geoffrey~E Hinton}.}
  \bibinfo{year}{2016}\natexlab{}.
\newblock \showarticletitle{Layer normalization}.
\newblock \bibinfo{journal}{\emph{arXiv preprint arXiv:1607.06450}}
  (\bibinfo{year}{2016}).
\newblock


\bibitem[Bahdanau et~al\mbox{.}(2014)]%
        {rnn2}
\bibfield{author}{\bibinfo{person}{Dzmitry Bahdanau},
  \bibinfo{person}{Kyunghyun Cho}, {and} \bibinfo{person}{Yoshua Bengio}.}
  \bibinfo{year}{2014}\natexlab{}.
\newblock \showarticletitle{Neural machine translation by jointly learning to
  align and translate}.
\newblock \bibinfo{journal}{\emph{arXiv preprint arXiv:1409.0473}}
  (\bibinfo{year}{2014}).
\newblock


\bibitem[Bai et~al\mbox{.}(2020)]%
        {tn4}
\bibfield{author}{\bibinfo{person}{Lei Bai}, \bibinfo{person}{Lina Yao},
  \bibinfo{person}{Can Li}, \bibinfo{person}{Xianzhi Wang}, {and}
  \bibinfo{person}{Can Wang}.} \bibinfo{year}{2020}\natexlab{}.
\newblock \showarticletitle{Adaptive graph convolutional recurrent network for
  traffic forecasting}.
\newblock \bibinfo{journal}{\emph{Advances in neural information processing
  systems}}  \bibinfo{volume}{33} (\bibinfo{year}{2020}),
  \bibinfo{pages}{17804--17815}.
\newblock


\bibitem[Barros et~al\mbox{.}(2021)]%
        {barros2021survey}
\bibfield{author}{\bibinfo{person}{Claudio~DT Barros},
  \bibinfo{person}{Matheus~RF Mendon{\c{c}}a}, \bibinfo{person}{Alex~B Vieira},
  {and} \bibinfo{person}{Artur Ziviani}.} \bibinfo{year}{2021}\natexlab{}.
\newblock \showarticletitle{A survey on embedding dynamic graphs}.
\newblock \bibinfo{journal}{\emph{ACM Computing Surveys (CSUR)}}
  \bibinfo{volume}{55}, \bibinfo{number}{1} (\bibinfo{year}{2021}),
  \bibinfo{pages}{1--37}.
\newblock


\bibitem[Battaglia et~al\mbox{.}(2018)]%
        {battaglia2018relational}
\bibfield{author}{\bibinfo{person}{Peter~W Battaglia},
  \bibinfo{person}{Jessica~B Hamrick}, \bibinfo{person}{Victor Bapst},
  \bibinfo{person}{Alvaro Sanchez-Gonzalez}, \bibinfo{person}{Vinicius
  Zambaldi}, \bibinfo{person}{Mateusz Malinowski}, \bibinfo{person}{Andrea
  Tacchetti}, \bibinfo{person}{David Raposo}, \bibinfo{person}{Adam Santoro},
  \bibinfo{person}{Ryan Faulkner}, {et~al\mbox{.}}}
  \bibinfo{year}{2018}\natexlab{}.
\newblock \showarticletitle{Relational inductive biases, deep learning, and
  graph networks}.
\newblock \bibinfo{journal}{\emph{arXiv preprint arXiv:1806.01261}}
  (\bibinfo{year}{2018}).
\newblock


\bibitem[Battiston et~al\mbox{.}(2020)]%
        {battiston2020networks}
\bibfield{author}{\bibinfo{person}{Federico Battiston}, \bibinfo{person}{Giulia
  Cencetti}, \bibinfo{person}{Iacopo Iacopini}, \bibinfo{person}{Vito Latora},
  \bibinfo{person}{Maxime Lucas}, \bibinfo{person}{Alice Patania},
  \bibinfo{person}{Jean-Gabriel Young}, {and} \bibinfo{person}{Giovanni
  Petri}.} \bibinfo{year}{2020}\natexlab{}.
\newblock \showarticletitle{Networks beyond pairwise interactions: structure
  and dynamics}.
\newblock \bibinfo{journal}{\emph{Physics Reports}}  \bibinfo{volume}{874}
  (\bibinfo{year}{2020}), \bibinfo{pages}{1--92}.
\newblock


\bibitem[Beltagy et~al\mbox{.}(2020)]%
        {swa}
\bibfield{author}{\bibinfo{person}{Iz Beltagy}, \bibinfo{person}{Matthew~E
  Peters}, {and} \bibinfo{person}{Arman Cohan}.}
  \bibinfo{year}{2020}\natexlab{}.
\newblock \showarticletitle{Longformer: The long-document transformer}.
\newblock \bibinfo{journal}{\emph{arXiv preprint arXiv:2004.05150}}
  (\bibinfo{year}{2020}).
\newblock


\bibitem[Benson et~al\mbox{.}(2018)]%
        {benson2018simplicial}
\bibfield{author}{\bibinfo{person}{Austin~R Benson} {et~al\mbox{.}}}
  \bibinfo{year}{2018}\natexlab{}.
\newblock \showarticletitle{Simplicial closure and higher-order link
  prediction}.
\newblock \bibinfo{journal}{\emph{Proceedings of the National Academy of
  Sciences}} \bibinfo{volume}{115}, \bibinfo{number}{48}
  (\bibinfo{year}{2018}), \bibinfo{pages}{E11221--E11230}.
\newblock


\bibitem[Besta(2021)]%
        {besta2021enabling}
\bibfield{author}{\bibinfo{person}{Maciej Besta}.}
  \bibinfo{year}{2021}\natexlab{}.
\newblock \emph{\bibinfo{title}{Enabling High-Performance Large-Scale Irregular
  Computations}}.
\newblock \bibinfo{thesistype}{Ph.\,D. Dissertation}. \bibinfo{school}{ETH
  Zurich}.
\newblock


\bibitem[Besta et~al\mbox{.}(2022a)]%
        {besta2019practice}
\bibfield{author}{\bibinfo{person}{Maciej Besta} {et~al\mbox{.}}}
  \bibinfo{year}{2022}\natexlab{a}.
\newblock \showarticletitle{Practice of Streaming Processing of Dynamic Graphs:
  Concepts, Models, and Systems}.
\newblock \bibinfo{journal}{\emph{IEEE TPDS}} (\bibinfo{year}{2022}).
\newblock


\bibitem[Besta et~al\mbox{.}(2023a)]%
        {besta2023thegdi}
\bibfield{author}{\bibinfo{person}{Maciej Besta} {et~al\mbox{.}}}
  \bibinfo{year}{2023}\natexlab{a}.
\newblock \showarticletitle{The Graph Database Interface: Scaling Online
  Transactional and Analytical Graph Workloads to Hundreds of Thousands of
  Cores}. In \bibinfo{booktitle}{\emph{ACM/IEEE Supercomputing}}.
\newblock


\bibitem[Besta et~al\mbox{.}(2023b)]%
        {besta2023gdi}
\bibfield{author}{\bibinfo{person}{Maciej Besta}, \bibinfo{person}{Robert
  Gerstenberger}, \bibinfo{person}{Nils Blach}, \bibinfo{person}{Marc Fischer},
  {and} \bibinfo{person}{Torsten Hoefler}.} \bibinfo{year}{2023}\natexlab{b}.
\newblock \bibinfo{booktitle}{\emph{GDI: A Graph Database Interface Standard}}.
\newblock \bibinfo{type}{{T}echnical {R}eport}.
\newblock
\newblock
\shownote{Available at
  \url{https://spcl.inf.ethz.ch/Research/Parallel_Programming/GDI/}}.


\bibitem[Besta et~al\mbox{.}(2023c)]%
        {besta2019demystifying}
\bibfield{author}{\bibinfo{person}{Maciej Besta}, \bibinfo{person}{Robert
  Gerstenberger}, \bibinfo{person}{Peter Emanuel}, \bibinfo{person}{Marc
  Fischer}, \bibinfo{person}{Micha{\l} Podstawski}, \bibinfo{person}{Claude
  Barthels}, \bibinfo{person}{Gustavo Alonso}, {and} \bibinfo{person}{Torsten
  Hoefler}.} \bibinfo{year}{2023}\natexlab{c}.
\newblock \showarticletitle{Demystifying Graph Databases: Analysis and Taxonomy
  of Data Organization, System Designs, and Graph Queries}.
\newblock \bibinfo{journal}{\emph{ACM CSUR}} (\bibinfo{year}{2023}).
\newblock


\bibitem[Besta et~al\mbox{.}(2022b)]%
        {besta2021motif}
\bibfield{author}{\bibinfo{person}{Maciej Besta}, \bibinfo{person}{Raphael
  Grob}, \bibinfo{person}{Cesare Miglioli}, \bibinfo{person}{Nicola Bernold},
  \bibinfo{person}{Grzegorz Kwasniewski}, \bibinfo{person}{Gabriel Gjini},
  \bibinfo{person}{Raghavendra Kanakagiri}, \bibinfo{person}{Saleh Ashkboos},
  \bibinfo{person}{Lukas Gianinazzi}, \bibinfo{person}{Nikoli Dryden},
  {et~al\mbox{.}}} \bibinfo{year}{2022}\natexlab{b}.
\newblock \showarticletitle{Motif Prediction with Graph Neural Networks}. In
  \bibinfo{booktitle}{\emph{ACM KDD}}.
\newblock


\bibitem[Besta and Hoefler(2023)]%
        {besta2022parallel}
\bibfield{author}{\bibinfo{person}{Maciej Besta} {and} \bibinfo{person}{Torsten
  Hoefler}.} \bibinfo{year}{2023}\natexlab{}.
\newblock \showarticletitle{Parallel and Distributed Graph Neural Networks: An
  In-Depth Concurrency Analysis}.
\newblock \bibinfo{journal}{\emph{IEEE TPAMI}} (\bibinfo{year}{2023}).
\newblock


\bibitem[Besta et~al\mbox{.}(2022c)]%
        {besta2022neural}
\bibfield{author}{\bibinfo{person}{Maciej Besta}, \bibinfo{person}{Patrick
  Iff}, \bibinfo{person}{Florian Scheidl}, \bibinfo{person}{Kazuki Osawa},
  \bibinfo{person}{Nikoli Dryden}, \bibinfo{person}{Michal Podstawski},
  \bibinfo{person}{Tiancheng Chen}, {and} \bibinfo{person}{Torsten Hoefler}.}
  \bibinfo{year}{2022}\natexlab{c}.
\newblock \showarticletitle{Neural Graph Databases}. In
  \bibinfo{booktitle}{\emph{LOG}}.
\newblock


\bibitem[Besta et~al\mbox{.}(2022d)]%
        {besta2022probgraph}
\bibfield{author}{\bibinfo{person}{Maciej Besta}, \bibinfo{person}{Cesare
  Miglioli}, \bibinfo{person}{Paolo~Sylos Labini}, \bibinfo{person}{Jakub
  T{\v{e}}tek}, \bibinfo{person}{Patrick Iff}, \bibinfo{person}{Raghavendra
  Kanakagiri}, \bibinfo{person}{Saleh Ashkboos}, \bibinfo{person}{Kacper
  Janda}, \bibinfo{person}{Micha{\l} Podstawski}, \bibinfo{person}{Grzegorz
  Kwa{\'s}niewski}, {et~al\mbox{.}}} \bibinfo{year}{2022}\natexlab{d}.
\newblock \showarticletitle{Probgraph: High-performance and high-accuracy graph
  mining with probabilistic set representations}. In
  \bibinfo{booktitle}{\emph{SC22: International Conference for High Performance
  Computing, Networking, Storage and Analysis}}. IEEE, \bibinfo{pages}{1--17}.
\newblock


\bibitem[Besta et~al\mbox{.}(2023d)]%
        {besta2023high}
\bibfield{author}{\bibinfo{person}{Maciej Besta}, \bibinfo{person}{Pawel Renc},
  \bibinfo{person}{Robert Gerstenberger}, \bibinfo{person}{Paolo Sylos~Labini},
  \bibinfo{person}{Alexandros Ziogas}, \bibinfo{person}{Tiancheng Chen},
  \bibinfo{person}{Lukas Gianinazzi}, \bibinfo{person}{Florian Scheidl},
  \bibinfo{person}{Kalman Szenes}, \bibinfo{person}{Armon Carigiet},
  {et~al\mbox{.}}} \bibinfo{year}{2023}\natexlab{d}.
\newblock \showarticletitle{High-Performance and Programmable Attentional Graph
  Neural Networks with Global Tensor Formulations}. In
  \bibinfo{booktitle}{\emph{Proceedings of the International Conference for
  High Performance Computing, Networking, Storage and Analysis}}.
  \bibinfo{pages}{1--16}.
\newblock


\bibitem[Besta et~al\mbox{.}(2021)]%
        {besta2021graphminesuite}
\bibfield{author}{\bibinfo{person}{Maciej Besta}, \bibinfo{person}{Zur
  Vonarburg-Shmaria}, \bibinfo{person}{Yannick Schaffner},
  \bibinfo{person}{Leonardo Schwarz}, \bibinfo{person}{Grzegorz Kwasniewski},
  \bibinfo{person}{Lukas Gianinazzi}, \bibinfo{person}{Jakub Beranek},
  \bibinfo{person}{Kacper Janda}, \bibinfo{person}{Tobias Holenstein},
  \bibinfo{person}{Sebastian Leisinger}, {et~al\mbox{.}}}
  \bibinfo{year}{2021}\natexlab{}.
\newblock \showarticletitle{Graphminesuite: Enabling high-performance and
  programmable graph mining algorithms with set algebra}.
\newblock \bibinfo{journal}{\emph{arXiv preprint arXiv:2103.03653}}
  (\bibinfo{year}{2021}).
\newblock


\bibitem[Bick et~al\mbox{.}(2021)]%
        {bick2021higher}
\bibfield{author}{\bibinfo{person}{Christian Bick}, \bibinfo{person}{Elizabeth
  Gross}, \bibinfo{person}{Heather~A Harrington}, {and}
  \bibinfo{person}{Michael~T Schaub}.} \bibinfo{year}{2021}\natexlab{}.
\newblock \showarticletitle{What are higher-order networks?}
\newblock \bibinfo{journal}{\emph{arXiv preprint arXiv:2104.11329}}
  (\bibinfo{year}{2021}).
\newblock


\bibitem[Bodnar et~al\mbox{.}(2021)]%
        {bodnar2021weisfeiler}
\bibfield{author}{\bibinfo{person}{Cristian Bodnar}, \bibinfo{person}{Fabrizio
  Frasca}, \bibinfo{person}{Nina Otter}, \bibinfo{person}{Yuguang Wang},
  \bibinfo{person}{Pietro Lio}, \bibinfo{person}{Guido~F Montufar}, {and}
  \bibinfo{person}{Michael Bronstein}.} \bibinfo{year}{2021}\natexlab{}.
\newblock \showarticletitle{Weisfeiler and lehman go cellular: Cw networks}.
\newblock \bibinfo{journal}{\emph{Advances in Neural Information Processing
  Systems}}  \bibinfo{volume}{34} (\bibinfo{year}{2021}),
  \bibinfo{pages}{2625--2640}.
\newblock


\bibitem[Bordes et~al\mbox{.}(2015)]%
        {mn1}
\bibfield{author}{\bibinfo{person}{Antoine Bordes}, \bibinfo{person}{Nicolas
  Usunier}, \bibinfo{person}{Sumit Chopra}, {and} \bibinfo{person}{Jason
  Weston}.} \bibinfo{year}{2015}\natexlab{}.
\newblock \bibinfo{title}{Large-scale Simple Question Answering with Memory
  Networks}.
\newblock
\newblock
\showeprint[arxiv]{1506.02075}~[cs.LG]


\bibitem[Bresson and Laurent(2017)]%
        {bresson2017residual}
\bibfield{author}{\bibinfo{person}{Xavier Bresson} {and}
  \bibinfo{person}{Thomas Laurent}.} \bibinfo{year}{2017}\natexlab{}.
\newblock \showarticletitle{Residual gated graph convnets}.
\newblock \bibinfo{journal}{\emph{arXiv preprint arXiv:1711.07553}}
  (\bibinfo{year}{2017}).
\newblock


\bibitem[Britz et~al\mbox{.}(2017)]%
        {rnn3}
\bibfield{author}{\bibinfo{person}{Denny Britz}, \bibinfo{person}{Anna Goldie},
  \bibinfo{person}{Minh-Thang Luong}, {and} \bibinfo{person}{Quoc Le}.}
  \bibinfo{year}{2017}\natexlab{}.
\newblock \showarticletitle{Massive exploration of neural machine translation
  architectures}.
\newblock \bibinfo{journal}{\emph{arXiv preprint arXiv:1703.03906}}
  (\bibinfo{year}{2017}).
\newblock


\bibitem[Bronstein et~al\mbox{.}(2017)]%
        {bronstein2017geometric}
\bibfield{author}{\bibinfo{person}{Michael~M Bronstein}, \bibinfo{person}{Joan
  Bruna}, \bibinfo{person}{Yann LeCun}, \bibinfo{person}{Arthur Szlam}, {and}
  \bibinfo{person}{Pierre Vandergheynst}.} \bibinfo{year}{2017}\natexlab{}.
\newblock \showarticletitle{Geometric deep learning: going beyond euclidean
  data}.
\newblock \bibinfo{journal}{\emph{IEEE Signal Processing Magazine}}
  \bibinfo{volume}{34}, \bibinfo{number}{4} (\bibinfo{year}{2017}),
  \bibinfo{pages}{18--42}.
\newblock


\bibitem[Brown et~al\mbox{.}(2020)]%
        {gpt3}
\bibfield{author}{\bibinfo{person}{Tom Brown}, \bibinfo{person}{Benjamin Mann},
  \bibinfo{person}{Nick Ryder}, \bibinfo{person}{Melanie Subbiah},
  \bibinfo{person}{Jared~D Kaplan}, \bibinfo{person}{Prafulla Dhariwal},
  \bibinfo{person}{Arvind Neelakantan}, \bibinfo{person}{Pranav Shyam},
  \bibinfo{person}{Girish Sastry}, \bibinfo{person}{Amanda Askell},
  {et~al\mbox{.}}} \bibinfo{year}{2020}\natexlab{}.
\newblock \showarticletitle{Language models are few-shot learners}.
\newblock \bibinfo{journal}{\emph{Advances in neural information processing
  systems}}  \bibinfo{volume}{33} (\bibinfo{year}{2020}),
  \bibinfo{pages}{1877--1901}.
\newblock


\bibitem[Cao et~al\mbox{.}(2021)]%
        {cao2021hift}
\bibfield{author}{\bibinfo{person}{Ziang Cao}, \bibinfo{person}{Changhong Fu},
  \bibinfo{person}{Junjie Ye}, \bibinfo{person}{Bowen Li}, {and}
  \bibinfo{person}{Yiming Li}.} \bibinfo{year}{2021}\natexlab{}.
\newblock \showarticletitle{Hift: Hierarchical feature transformer for aerial
  tracking}. In \bibinfo{booktitle}{\emph{Proceedings of the IEEE/CVF
  International Conference on Computer Vision}}. \bibinfo{pages}{15457--15466}.
\newblock


\bibitem[Chami et~al\mbox{.}(2020)]%
        {chami2020machine}
\bibfield{author}{\bibinfo{person}{Ines Chami}, \bibinfo{person}{Sami
  {Abu-El-Haija}}, \bibinfo{person}{Bryan Perozzi},
  \bibinfo{person}{Christopher R{\'e}}, {and} \bibinfo{person}{Kevin Murphy}.}
  \bibinfo{year}{2020}\natexlab{}.
\newblock \showarticletitle{Machine learning on graphs: A model and
  comprehensive taxonomy}.
\newblock \bibinfo{journal}{\emph{arXiv preprint arXiv:2005.03675}}
  (\bibinfo{year}{2020}).
\newblock


\bibitem[Chang et~al\mbox{.}(2020)]%
        {con6}
\bibfield{author}{\bibinfo{person}{Xiaofu Chang}, \bibinfo{person}{Xuqin Liu},
  \bibinfo{person}{Jianfeng Wen}, \bibinfo{person}{Shuang Li},
  \bibinfo{person}{Yanming Fang}, \bibinfo{person}{Le Song}, {and}
  \bibinfo{person}{Yuan Qi}.} \bibinfo{year}{2020}\natexlab{}.
\newblock \showarticletitle{Continuous-time dynamic graph learning via neural
  interaction processes}. In \bibinfo{booktitle}{\emph{Proceedings of the 29th
  ACM International Conference on Information \& Knowledge Management}}.
  \bibinfo{pages}{145--154}.
\newblock


\bibitem[Choudhury et~al\mbox{.}(2017)]%
        {choudhury2017nous}
\bibfield{author}{\bibinfo{person}{Sutanay Choudhury}, \bibinfo{person}{Khushbu
  Agarwal}, \bibinfo{person}{Sumit Purohit}, \bibinfo{person}{Baichuan Zhang},
  \bibinfo{person}{Meg Pirrung}, \bibinfo{person}{Will Smith}, {and}
  \bibinfo{person}{Mathew Thomas}.} \bibinfo{year}{2017}\natexlab{}.
\newblock \showarticletitle{Nous: Construction and querying of dynamic
  knowledge graphs}. In \bibinfo{booktitle}{\emph{IEEE ICDE}}.
  \bibinfo{pages}{1563--1565}.
\newblock


\bibitem[Cong et~al\mbox{.}(2023)]%
        {graphmixer}
\bibfield{author}{\bibinfo{person}{Weilin Cong}, \bibinfo{person}{Si Zhang},
  \bibinfo{person}{Jian Kang}, \bibinfo{person}{Baichuan Yuan},
  \bibinfo{person}{Hao Wu}, \bibinfo{person}{Xin Zhou},
  \bibinfo{person}{Hanghang Tong}, {and} \bibinfo{person}{Mehrdad Mahdavi}.}
  \bibinfo{year}{2023}\natexlab{}.
\newblock \showarticletitle{Do We Really Need Complicated Model Architectures
  For Temporal Networks?}
\newblock \bibinfo{journal}{\emph{arXiv preprint arXiv:2302.11636}}
  (\bibinfo{year}{2023}).
\newblock


\bibitem[Dai et~al\mbox{.}(2019)]%
        {transxl}
\bibfield{author}{\bibinfo{person}{Zihang Dai}, \bibinfo{person}{Zhilin Yang},
  \bibinfo{person}{Yiming Yang}, \bibinfo{person}{Jaime Carbonell},
  \bibinfo{person}{Quoc~V Le}, {and} \bibinfo{person}{Ruslan Salakhutdinov}.}
  \bibinfo{year}{2019}\natexlab{}.
\newblock \showarticletitle{Transformer-xl: Attentive language models beyond a
  fixed-length context}.
\newblock \bibinfo{journal}{\emph{arXiv preprint arXiv:1901.02860}}
  (\bibinfo{year}{2019}).
\newblock


\bibitem[Dao et~al\mbox{.}(2022)]%
        {flashatt}
\bibfield{author}{\bibinfo{person}{Tri Dao}, \bibinfo{person}{Dan Fu},
  \bibinfo{person}{Stefano Ermon}, \bibinfo{person}{Atri Rudra}, {and}
  \bibinfo{person}{Christopher R{\'e}}.} \bibinfo{year}{2022}\natexlab{}.
\newblock \showarticletitle{Flashattention: Fast and memory-efficient exact
  attention with io-awareness}.
\newblock \bibinfo{journal}{\emph{Advances in Neural Information Processing
  Systems}}  \bibinfo{volume}{35} (\bibinfo{year}{2022}),
  \bibinfo{pages}{16344--16359}.
\newblock


\bibitem[Davoudian et~al\mbox{.}(2018)]%
        {davoudian2018survey}
\bibfield{author}{\bibinfo{person}{Ali Davoudian}, \bibinfo{person}{Liu Chen},
  {and} \bibinfo{person}{Mengchi Liu}.} \bibinfo{year}{2018}\natexlab{}.
\newblock \showarticletitle{{A survey on NoSQL stores}}.
\newblock \bibinfo{journal}{\emph{ACM Computing Surveys (CSUR)}}
  \bibinfo{volume}{51}, \bibinfo{number}{2}, Article \bibinfo{articleno}{40}
  (\bibinfo{year}{2018}), \bibinfo{numpages}{43}~pages.
\newblock
\showISSN{0360-0300}
\urldef\tempurl%
\url{https://doi.org/10.1145/3158661}
\showDOI{\tempurl}


\bibitem[Dhulipala et~al\mbox{.}(2019)]%
        {dhulipala2019low}
\bibfield{author}{\bibinfo{person}{Laxman Dhulipala} {et~al\mbox{.}}}
  \bibinfo{year}{2019}\natexlab{}.
\newblock \showarticletitle{Low-Latency Graph Streaming Using Compressed
  Purely-Functional Trees}.
\newblock \bibinfo{journal}{\emph{arXiv:1904.08380}} (\bibinfo{year}{2019}).
\newblock


\bibitem[Dosovitskiy et~al\mbox{.}(2020)]%
        {vit}
\bibfield{author}{\bibinfo{person}{Alexey Dosovitskiy}, \bibinfo{person}{Lucas
  Beyer}, \bibinfo{person}{Alexander Kolesnikov}, \bibinfo{person}{Dirk
  Weissenborn}, \bibinfo{person}{Xiaohua Zhai}, \bibinfo{person}{Thomas
  Unterthiner}, \bibinfo{person}{Mostafa Dehghani}, \bibinfo{person}{Matthias
  Minderer}, \bibinfo{person}{Georg Heigold}, \bibinfo{person}{Sylvain Gelly},
  {et~al\mbox{.}}} \bibinfo{year}{2020}\natexlab{}.
\newblock \showarticletitle{An image is worth 16x16 words: Transformers for
  image recognition at scale}.
\newblock \bibinfo{journal}{\emph{arXiv preprint arXiv:2010.11929}}
  (\bibinfo{year}{2020}).
\newblock


\bibitem[Fan et~al\mbox{.}(2021)]%
        {ui2}
\bibfield{author}{\bibinfo{person}{Ziwei Fan}, \bibinfo{person}{Zhiwei Liu},
  \bibinfo{person}{Jiawei Zhang}, \bibinfo{person}{Yun Xiong},
  \bibinfo{person}{Lei Zheng}, {and} \bibinfo{person}{Philip~S Yu}.}
  \bibinfo{year}{2021}\natexlab{}.
\newblock \showarticletitle{Continuous-time sequential recommendation with
  temporal graph collaborative transformer}. In
  \bibinfo{booktitle}{\emph{Proceedings of the 30th ACM international
  conference on information \& knowledge management}}.
  \bibinfo{pages}{433--442}.
\newblock


\bibitem[Feng et~al\mbox{.}(2015)]%
        {feng2015distinger}
\bibfield{author}{\bibinfo{person}{Guoyao Feng} {et~al\mbox{.}}}
  \bibinfo{year}{2015}\natexlab{}.
\newblock \showarticletitle{DISTINGER: A distributed graph data structure for
  massive dynamic graph processing}. In \bibinfo{booktitle}{\emph{IEEE Big
  Data}}. \bibinfo{pages}{1814--1822}.
\newblock


\bibitem[Fey and Lenssen(2019)]%
        {fey2019fast}
\bibfield{author}{\bibinfo{person}{Matthias Fey} {and}
  \bibinfo{person}{Jan~Eric Lenssen}.} \bibinfo{year}{2019}\natexlab{}.
\newblock \showarticletitle{Fast graph representation learning with PyTorch
  Geometric}.
\newblock \bibinfo{journal}{\emph{arXiv preprint arXiv:1903.02428}}
  (\bibinfo{year}{2019}).
\newblock


\bibitem[Frasca et~al\mbox{.}(2022)]%
        {frasca2022understanding}
\bibfield{author}{\bibinfo{person}{Fabrizio Frasca}, \bibinfo{person}{Beatrice
  Bevilacqua}, \bibinfo{person}{Michael Bronstein}, {and}
  \bibinfo{person}{Haggai Maron}.} \bibinfo{year}{2022}\natexlab{}.
\newblock \showarticletitle{Understanding and extending subgraph gnns by
  rethinking their symmetries}.
\newblock \bibinfo{journal}{\emph{Advances in Neural Information Processing
  Systems}}  \bibinfo{volume}{35} (\bibinfo{year}{2022}),
  \bibinfo{pages}{31376--31390}.
\newblock


\bibitem[Frasca et~al\mbox{.}(2020)]%
        {frasca2020sign}
\bibfield{author}{\bibinfo{person}{Fabrizio Frasca}, \bibinfo{person}{Emanuele
  Rossi}, \bibinfo{person}{Davide Eynard}, \bibinfo{person}{Ben Chamberlain},
  \bibinfo{person}{Michael Bronstein}, {and} \bibinfo{person}{Federico Monti}.}
  \bibinfo{year}{2020}\natexlab{}.
\newblock \showarticletitle{Sign: Scalable inception graph neural networks}.
\newblock \bibinfo{journal}{\emph{arXiv preprint arXiv:2004.11198}}
  (\bibinfo{year}{2020}).
\newblock


\bibitem[Gajendran(2012)]%
        {gajendran2012survey}
\bibfield{author}{\bibinfo{person}{Santhosh~Kumar Gajendran}.}
  \bibinfo{year}{2012}\natexlab{}.
\newblock \showarticletitle{{A survey on NoSQL databases}}.
\newblock \bibinfo{journal}{\emph{University of Illinois}}
  (\bibinfo{year}{2012}).
\newblock


\bibitem[Geng et~al\mbox{.}(2020)]%
        {geng2020awb}
\bibfield{author}{\bibinfo{person}{Tong Geng}, \bibinfo{person}{Ang Li},
  \bibinfo{person}{Runbin Shi}, \bibinfo{person}{Chunshu Wu},
  \bibinfo{person}{Tianqi Wang}, \bibinfo{person}{Yanfei Li},
  \bibinfo{person}{Pouya Haghi}, \bibinfo{person}{Antonino Tumeo},
  \bibinfo{person}{Shuai Che}, \bibinfo{person}{Steve Reinhardt},
  {et~al\mbox{.}}} \bibinfo{year}{2020}\natexlab{}.
\newblock \showarticletitle{AWB-GCN: A graph convolutional network accelerator
  with runtime workload rebalancing}. In \bibinfo{booktitle}{\emph{IEEE/ACM
  MICRO}}.
\newblock


\bibitem[Gianinazzi et~al\mbox{.}(2021)]%
        {gianinazzi2021learning}
\bibfield{author}{\bibinfo{person}{Lukas Gianinazzi},
  \bibinfo{person}{Maximilian Fries}, \bibinfo{person}{Nikoli Dryden},
  \bibinfo{person}{Tal Ben-Nun}, {and} \bibinfo{person}{Torsten Hoefler}.}
  \bibinfo{year}{2021}\natexlab{}.
\newblock \showarticletitle{Learning Combinatorial Node Labeling Algorithms}.
\newblock \bibinfo{journal}{\emph{arXiv preprint arXiv:2106.03594}}
  (\bibinfo{year}{2021}).
\newblock


\bibitem[Gilmer et~al\mbox{.}(2017)]%
        {gilmer2017neural}
\bibfield{author}{\bibinfo{person}{Justin Gilmer}, \bibinfo{person}{Samuel~S
  Schoenholz}, \bibinfo{person}{Patrick~F Riley}, \bibinfo{person}{Oriol
  Vinyals}, {and} \bibinfo{person}{George~E Dahl}.}
  \bibinfo{year}{2017}\natexlab{}.
\newblock \showarticletitle{Neural message passing for quantum chemistry}. In
  \bibinfo{booktitle}{\emph{International Conference on Machine Learning}}.
  PMLR, \bibinfo{pages}{1263--1272}.
\newblock


\bibitem[Guo et~al\mbox{.}(2019)]%
        {tn3}
\bibfield{author}{\bibinfo{person}{Shengnan Guo}, \bibinfo{person}{Youfang
  Lin}, \bibinfo{person}{Ning Feng}, \bibinfo{person}{Chao Song}, {and}
  \bibinfo{person}{Huaiyu Wan}.} \bibinfo{year}{2019}\natexlab{}.
\newblock \showarticletitle{Attention based spatial-temporal graph
  convolutional networks for traffic flow forecasting}. In
  \bibinfo{booktitle}{\emph{Proceedings of the AAAI conference on artificial
  intelligence}}, Vol.~\bibinfo{volume}{33}. \bibinfo{pages}{922--929}.
\newblock


\bibitem[Hamilton et~al\mbox{.}(2017a)]%
        {hamilton2017representation}
\bibfield{author}{\bibinfo{person}{William~L Hamilton} {et~al\mbox{.}}}
  \bibinfo{year}{2017}\natexlab{a}.
\newblock \showarticletitle{Representation learning on graphs: Methods and
  applications}.
\newblock \bibinfo{journal}{\emph{arXiv preprint arXiv:1709.05584}}
  (\bibinfo{year}{2017}).
\newblock


\bibitem[Hamilton et~al\mbox{.}(2017b)]%
        {hamilton2017inductive}
\bibfield{author}{\bibinfo{person}{William~L Hamilton}, \bibinfo{person}{Rex
  Ying}, {and} \bibinfo{person}{Jure Leskovec}.}
  \bibinfo{year}{2017}\natexlab{b}.
\newblock \showarticletitle{Inductive representation learning on large graphs}.
  In \bibinfo{booktitle}{\emph{NeurIPS}}.
\newblock


\bibitem[Han et~al\mbox{.}(2011)]%
        {han2011survey}
\bibfield{author}{\bibinfo{person}{Jing Han}, \bibinfo{person}{E Haihong},
  \bibinfo{person}{Guan Le}, {and} \bibinfo{person}{Jian Du}.}
  \bibinfo{year}{2011}\natexlab{}.
\newblock \showarticletitle{{Survey on NoSQL database}}. In
  \bibinfo{booktitle}{\emph{2011 6th international conference on pervasive
  computing and applications}}. IEEE, \bibinfo{pages}{363--366}.
\newblock


\bibitem[Han et~al\mbox{.}(2021)]%
        {han2021dynamic}
\bibfield{author}{\bibinfo{person}{Yizeng Han}, \bibinfo{person}{Gao Huang},
  \bibinfo{person}{Shiji Song}, \bibinfo{person}{Le Yang},
  \bibinfo{person}{Honghui Wang}, {and} \bibinfo{person}{Yulin Wang}.}
  \bibinfo{year}{2021}\natexlab{}.
\newblock \showarticletitle{Dynamic neural networks: A survey}.
\newblock \bibinfo{journal}{\emph{IEEE Transactions on Pattern Analysis and
  Machine Intelligence}} \bibinfo{volume}{44}, \bibinfo{number}{11}
  (\bibinfo{year}{2021}), \bibinfo{pages}{7436--7456}.
\newblock


\bibitem[He et~al\mbox{.}(2016)]%
        {rc}
\bibfield{author}{\bibinfo{person}{Kaiming He}, \bibinfo{person}{Xiangyu
  Zhang}, \bibinfo{person}{Shaoqing Ren}, {and} \bibinfo{person}{Jian Sun}.}
  \bibinfo{year}{2016}\natexlab{}.
\newblock \showarticletitle{Deep residual learning for image recognition}. In
  \bibinfo{booktitle}{\emph{Proceedings of the IEEE conference on computer
  vision and pattern recognition}}. \bibinfo{pages}{770--778}.
\newblock


\bibitem[Hochreiter and Schmidhuber(1997)]%
        {lstm}
\bibfield{author}{\bibinfo{person}{Sepp Hochreiter} {and}
  \bibinfo{person}{J\"{u}rgen Schmidhuber}.} \bibinfo{year}{1997}\natexlab{}.
\newblock \showarticletitle{Long Short-Term Memory}.
\newblock  \bibinfo{volume}{9}, \bibinfo{number}{8} (\bibinfo{year}{1997}),
  \bibinfo{numpages}{46}~pages.
\newblock
\showISSN{0899-7667}
\urldef\tempurl%
\url{https://doi.org/10.1162/neco.1997.9.8.1735}
\showDOI{\tempurl}


\bibitem[Hu et~al\mbox{.}(2020)]%
        {hu2020featgraph}
\bibfield{author}{\bibinfo{person}{Yuwei Hu} {et~al\mbox{.}}}
  \bibinfo{year}{2020}\natexlab{}.
\newblock \showarticletitle{Featgraph: A flexible and efficient backend for
  graph neural network systems}.
\newblock \bibinfo{journal}{\emph{arXiv preprint arXiv:2008.11359}}
  (\bibinfo{year}{2020}).
\newblock


\bibitem[Huang et~al\mbox{.}(2020)]%
        {ps1}
\bibfield{author}{\bibinfo{person}{Zijie Huang}, \bibinfo{person}{Yizhou Sun},
  {and} \bibinfo{person}{Wei Wang}.} \bibinfo{year}{2020}\natexlab{}.
\newblock \showarticletitle{Learning continuous system dynamics from
  irregularly-sampled partial observations}.
\newblock \bibinfo{journal}{\emph{Advances in Neural Information Processing
  Systems}}  \bibinfo{volume}{33} (\bibinfo{year}{2020}),
  \bibinfo{pages}{16177--16187}.
\newblock


\bibitem[Hutchins et~al\mbox{.}(2022)]%
        {brt}
\bibfield{author}{\bibinfo{person}{DeLesley Hutchins}, \bibinfo{person}{Imanol
  Schlag}, \bibinfo{person}{Yuhuai Wu}, \bibinfo{person}{Ethan Dyer}, {and}
  \bibinfo{person}{Behnam Neyshabur}.} \bibinfo{year}{2022}\natexlab{}.
\newblock \showarticletitle{Block-recurrent transformers}.
\newblock \bibinfo{journal}{\emph{arXiv preprint arXiv:2203.07852}}
  (\bibinfo{year}{2022}).
\newblock


\bibitem[Ivanov and Burnaev(2018)]%
        {aw}
\bibfield{author}{\bibinfo{person}{Sergey Ivanov} {and} \bibinfo{person}{Evgeny
  Burnaev}.} \bibinfo{year}{2018}\natexlab{}.
\newblock \showarticletitle{Anonymous walk embeddings}. In
  \bibinfo{booktitle}{\emph{International conference on machine learning}}.
  PMLR, \bibinfo{pages}{2186--2195}.
\newblock


\bibitem[Jia et~al\mbox{.}(2020)]%
        {jia2020improving}
\bibfield{author}{\bibinfo{person}{Zhihao Jia} {et~al\mbox{.}}}
  \bibinfo{year}{2020}\natexlab{}.
\newblock \showarticletitle{Improving the accuracy, scalability, and
  performance of graph neural networks with roc}.
\newblock \bibinfo{journal}{\emph{MLSys}} (\bibinfo{year}{2020}).
\newblock


\bibitem[Jin et~al\mbox{.}(2022)]%
        {con10}
\bibfield{author}{\bibinfo{person}{Ming Jin}, \bibinfo{person}{Yuan-Fang Li},
  {and} \bibinfo{person}{Shirui Pan}.} \bibinfo{year}{2022}\natexlab{}.
\newblock \showarticletitle{Neural temporal walks: Motif-aware representation
  learning on continuous-time dynamic graphs}.
\newblock \bibinfo{journal}{\emph{Advances in Neural Information Processing
  Systems}}  \bibinfo{volume}{35} (\bibinfo{year}{2022}),
  \bibinfo{pages}{19874--19886}.
\newblock


\bibitem[Kaliyar(2015)]%
        {gdb_survey_paper_Kaliyar}
\bibfield{author}{\bibinfo{person}{R.~Kumar Kaliyar}.}
  \bibinfo{year}{2015}\natexlab{}.
\newblock \showarticletitle{Graph databases: A survey}. In
  \bibinfo{booktitle}{\emph{ICCCA}}. \bibinfo{pages}{785--790}.
\newblock


\bibitem[Kazemi et~al\mbox{.}(2020a)]%
        {kazemi}
\bibfield{author}{\bibinfo{person}{Seyed~Mehran Kazemi},
  \bibinfo{person}{Rishab Goel}, \bibinfo{person}{Kshitij Jain},
  \bibinfo{person}{Ivan Kobyzev}, \bibinfo{person}{Akshay Sethi},
  \bibinfo{person}{Peter Forsyth}, {and} \bibinfo{person}{Pascal Poupart}.}
  \bibinfo{year}{2020}\natexlab{a}.
\newblock \showarticletitle{Representation learning for dynamic graphs: A
  survey}.
\newblock \bibinfo{journal}{\emph{The Journal of Machine Learning Research}}
  \bibinfo{volume}{21}, \bibinfo{number}{1} (\bibinfo{year}{2020}),
  \bibinfo{pages}{2648--2720}.
\newblock


\bibitem[Kazemi et~al\mbox{.}(2020b)]%
        {kazemi2020representation}
\bibfield{author}{\bibinfo{person}{Seyed~Mehran Kazemi},
  \bibinfo{person}{Rishab Goel}, \bibinfo{person}{Kshitij Jain},
  \bibinfo{person}{Ivan Kobyzev}, \bibinfo{person}{Akshay Sethi},
  \bibinfo{person}{Peter Forsyth}, {and} \bibinfo{person}{Pascal Poupart}.}
  \bibinfo{year}{2020}\natexlab{b}.
\newblock \showarticletitle{Representation learning for dynamic graphs: A
  survey}.
\newblock \bibinfo{journal}{\emph{The Journal of Machine Learning Research}}
  \bibinfo{volume}{21}, \bibinfo{number}{1} (\bibinfo{year}{2020}),
  \bibinfo{pages}{2648--2720}.
\newblock


\bibitem[Kingma and Ba(2014)]%
        {kingma2014adam}
\bibfield{author}{\bibinfo{person}{Diederik~P Kingma} {and}
  \bibinfo{person}{Jimmy Ba}.} \bibinfo{year}{2014}\natexlab{}.
\newblock \showarticletitle{Adam: A method for stochastic optimization}.
\newblock \bibinfo{journal}{\emph{arXiv preprint arXiv:1412.6980}}
  (\bibinfo{year}{2014}).
\newblock


\bibitem[Kiningham et~al\mbox{.}(2020a)]%
        {kiningham2020greta}
\bibfield{author}{\bibinfo{person}{Kevin Kiningham}, \bibinfo{person}{Philip
  Levis}, {and} \bibinfo{person}{Christopher R{\'e}}.}
  \bibinfo{year}{2020}\natexlab{a}.
\newblock \showarticletitle{GReTA: Hardware Optimized Graph Processing for
  GNNs}. In \bibinfo{booktitle}{\emph{ReCoML}}.
\newblock


\bibitem[Kiningham et~al\mbox{.}(2020b)]%
        {kiningham2020grip}
\bibfield{author}{\bibinfo{person}{Kevin Kiningham},
  \bibinfo{person}{Christopher Re}, {and} \bibinfo{person}{Philip Levis}.}
  \bibinfo{year}{2020}\natexlab{b}.
\newblock \showarticletitle{GRIP: a graph neural network accelerator
  architecture}.
\newblock \bibinfo{journal}{\emph{arXiv preprint arXiv:2007.13828}}
  (\bibinfo{year}{2020}).
\newblock


\bibitem[Kipf and Welling(2016)]%
        {kipf2016semi}
\bibfield{author}{\bibinfo{person}{Thomas~N Kipf} {and} \bibinfo{person}{Max
  Welling}.} \bibinfo{year}{2016}\natexlab{}.
\newblock \showarticletitle{Semi-supervised classification with graph
  convolutional networks}.
\newblock \bibinfo{journal}{\emph{arXiv preprint arXiv:1609.02907}}
  (\bibinfo{year}{2016}).
\newblock


\bibitem[Kumar et~al\mbox{.}(2019)]%
        {jodie}
\bibfield{author}{\bibinfo{person}{Srijan Kumar}, \bibinfo{person}{Xikun
  Zhang}, {and} \bibinfo{person}{Jure Leskovec}.}
  \bibinfo{year}{2019}\natexlab{}.
\newblock \showarticletitle{Predicting dynamic embedding trajectory in temporal
  interaction networks}. In \bibinfo{booktitle}{\emph{Proceedings of the 25th
  ACM SIGKDD international conference on knowledge discovery \& data mining}}.
  \bibinfo{pages}{1269--1278}.
\newblock


\bibitem[Kumar and Babu(2015)]%
        {kumar2015domain}
\bibfield{author}{\bibinfo{person}{Vijay Kumar} {and} \bibinfo{person}{Anjan
  Babu}.} \bibinfo{year}{2015}\natexlab{}.
\newblock \showarticletitle{Domain Suitable Graph Database Selection: A
  Preliminary Report}. In \bibinfo{booktitle}{\emph{3rd International
  Conference on Advances in Engineering Sciences \& Applied Mathematics,
  London, UK}}. \bibinfo{pages}{26--29}.
\newblock


\bibitem[Li et~al\mbox{.}(2020a)]%
        {li2020pytorch}
\bibfield{author}{\bibinfo{person}{Shen Li} {et~al\mbox{.}}}
  \bibinfo{year}{2020}\natexlab{a}.
\newblock \showarticletitle{Pytorch distributed: Experiences on accelerating
  data parallel training}.
\newblock \bibinfo{journal}{\emph{arXiv preprint arXiv:2006.15704}}
  (\bibinfo{year}{2020}).
\newblock


\bibitem[Li et~al\mbox{.}(2020b)]%
        {ui1}
\bibfield{author}{\bibinfo{person}{Xiaohan Li}, \bibinfo{person}{Mengqi Zhang},
  \bibinfo{person}{Shu Wu}, \bibinfo{person}{Zheng Liu}, \bibinfo{person}{Liang
  Wang}, {and} \bibinfo{person}{S~Yu Philip}.}
  \bibinfo{year}{2020}\natexlab{b}.
\newblock \showarticletitle{Dynamic graph collaborative filtering}. In
  \bibinfo{booktitle}{\emph{2020 IEEE international conference on data mining
  (ICDM)}}. IEEE, \bibinfo{pages}{322--331}.
\newblock


\bibitem[Liang et~al\mbox{.}(2020)]%
        {liang2020engn}
\bibfield{author}{\bibinfo{person}{Shengwen Liang}, \bibinfo{person}{Ying
  Wang}, \bibinfo{person}{Cheng Liu}, \bibinfo{person}{Lei He},
  \bibinfo{person}{LI Huawei}, \bibinfo{person}{Dawen Xu}, {and}
  \bibinfo{person}{Xiaowei Li}.} \bibinfo{year}{2020}\natexlab{}.
\newblock \showarticletitle{Engn: A high-throughput and energy-efficient
  accelerator for large graph neural networks}.
\newblock \bibinfo{journal}{\emph{IEEE TOC}} (\bibinfo{year}{2020}).
\newblock


\bibitem[Lim et~al\mbox{.}(2021)]%
        {tft}
\bibfield{author}{\bibinfo{person}{Bryan Lim}, \bibinfo{person}{Sercan~{\"O}
  Ar{\i}k}, \bibinfo{person}{Nicolas Loeff}, {and} \bibinfo{person}{Tomas
  Pfister}.} \bibinfo{year}{2021}\natexlab{}.
\newblock \showarticletitle{Temporal fusion transformers for interpretable
  multi-horizon time series forecasting}.
\newblock \bibinfo{journal}{\emph{International Journal of Forecasting}}
  \bibinfo{volume}{37}, \bibinfo{number}{4} (\bibinfo{year}{2021}),
  \bibinfo{pages}{1748--1764}.
\newblock


\bibitem[Liu et~al\mbox{.}(2019)]%
        {liu2019higher}
\bibfield{author}{\bibinfo{person}{Songtao Liu}, \bibinfo{person}{Lingwei
  Chen}, \bibinfo{person}{Hanze Dong}, \bibinfo{person}{Zihao Wang},
  \bibinfo{person}{Dinghao Wu}, {and} \bibinfo{person}{Zengfeng Huang}.}
  \bibinfo{year}{2019}\natexlab{}.
\newblock \showarticletitle{Higher-order weighted graph convolutional
  networks}.
\newblock \bibinfo{journal}{\emph{arXiv preprint arXiv:1911.04129}}
  (\bibinfo{year}{2019}).
\newblock


\bibitem[Liu et~al\mbox{.}(2021)]%
        {liu2021swin}
\bibfield{author}{\bibinfo{person}{Ze Liu}, \bibinfo{person}{Yutong Lin},
  \bibinfo{person}{Yue Cao}, \bibinfo{person}{Han Hu}, \bibinfo{person}{Yixuan
  Wei}, \bibinfo{person}{Zheng Zhang}, \bibinfo{person}{Stephen Lin}, {and}
  \bibinfo{person}{Baining Guo}.} \bibinfo{year}{2021}\natexlab{}.
\newblock \showarticletitle{Swin transformer: Hierarchical vision transformer
  using shifted windows}. In \bibinfo{booktitle}{\emph{Proceedings of the
  IEEE/CVF international conference on computer vision}}.
  \bibinfo{pages}{10012--10022}.
\newblock


\bibitem[Luo and Li(2022)]%
        {con11}
\bibfield{author}{\bibinfo{person}{Yuhong Luo} {and} \bibinfo{person}{Pan Li}.}
  \bibinfo{year}{2022}\natexlab{}.
\newblock \showarticletitle{Neighborhood-aware scalable temporal network
  representation learning}. In \bibinfo{booktitle}{\emph{Learning on Graphs
  Conference}}. PMLR, \bibinfo{pages}{1--1}.
\newblock


\bibitem[Ma et~al\mbox{.}(2019)]%
        {ma2019neugraph}
\bibfield{author}{\bibinfo{person}{Lingxiao Ma}, \bibinfo{person}{Zhi Yang},
  \bibinfo{person}{Youshan Miao}, \bibinfo{person}{Jilong Xue},
  \bibinfo{person}{Ming Wu}, \bibinfo{person}{Lidong Zhou}, {and}
  \bibinfo{person}{Yafei Dai}.} \bibinfo{year}{2019}\natexlab{}.
\newblock \showarticletitle{Neugraph: parallel deep neural network computation
  on large graphs}. In \bibinfo{booktitle}{\emph{USENIX ATC}}.
\newblock


\bibitem[Ma et~al\mbox{.}(2020)]%
        {con5}
\bibfield{author}{\bibinfo{person}{Yao Ma}, \bibinfo{person}{Ziyi Guo},
  \bibinfo{person}{Zhaocun Ren}, \bibinfo{person}{Jiliang Tang}, {and}
  \bibinfo{person}{Dawei Yin}.} \bibinfo{year}{2020}\natexlab{}.
\newblock \showarticletitle{Streaming graph neural networks}. In
  \bibinfo{booktitle}{\emph{Proceedings of the 43rd international ACM SIGIR
  conference on research and development in information retrieval}}.
  \bibinfo{pages}{719--728}.
\newblock


\bibitem[Monti et~al\mbox{.}(2017)]%
        {monti2017geometric}
\bibfield{author}{\bibinfo{person}{Federico Monti}, \bibinfo{person}{Davide
  Boscaini}, \bibinfo{person}{Jonathan Masci}, \bibinfo{person}{Emanuele
  Rodola}, \bibinfo{person}{Jan Svoboda}, {and} \bibinfo{person}{Michael~M
  Bronstein}.} \bibinfo{year}{2017}\natexlab{}.
\newblock \showarticletitle{Geometric deep learning on graphs and manifolds
  using mixture model cnns}. In \bibinfo{booktitle}{\emph{IEEE CVPR}}.
\newblock


\bibitem[Morris et~al\mbox{.}(2019)]%
        {morris2019weisfeiler}
\bibfield{author}{\bibinfo{person}{Christopher Morris}, \bibinfo{person}{Martin
  Ritzert}, \bibinfo{person}{Matthias Fey}, \bibinfo{person}{William~L
  Hamilton}, \bibinfo{person}{Jan~Eric Lenssen}, \bibinfo{person}{Gaurav
  Rattan}, {and} \bibinfo{person}{Martin Grohe}.}
  \bibinfo{year}{2019}\natexlab{}.
\newblock \showarticletitle{Weisfeiler and leman go neural: Higher-order graph
  neural networks}. In \bibinfo{booktitle}{\emph{Proceedings of the AAAI
  conference on artificial intelligence}}, Vol.~\bibinfo{volume}{33}.
  \bibinfo{pages}{4602--4609}.
\newblock


\bibitem[Pappagari et~al\mbox{.}(2019)]%
        {pappagari2019hierarchical}
\bibfield{author}{\bibinfo{person}{Raghavendra Pappagari},
  \bibinfo{person}{Piotr Zelasko}, \bibinfo{person}{Jes{\'u}s Villalba},
  \bibinfo{person}{Yishay Carmiel}, {and} \bibinfo{person}{Najim Dehak}.}
  \bibinfo{year}{2019}\natexlab{}.
\newblock \showarticletitle{Hierarchical transformers for long document
  classification}. In \bibinfo{booktitle}{\emph{2019 IEEE automatic speech
  recognition and understanding workshop (ASRU)}}. IEEE,
  \bibinfo{pages}{838--844}.
\newblock


\bibitem[Pedregosa et~al\mbox{.}(2011)]%
        {scikit}
\bibfield{author}{\bibinfo{person}{F. Pedregosa}, \bibinfo{person}{G.
  Varoquaux}, \bibinfo{person}{A. Gramfort}, \bibinfo{person}{V. Michel},
  \bibinfo{person}{B. Thirion}, \bibinfo{person}{O. Grisel},
  \bibinfo{person}{M. Blondel}, \bibinfo{person}{P. Prettenhofer},
  \bibinfo{person}{R. Weiss}, \bibinfo{person}{V. Dubourg}, \bibinfo{person}{J.
  Vanderplas}, \bibinfo{person}{A. Passos}, \bibinfo{person}{D. Cournapeau},
  \bibinfo{person}{M. Brucher}, \bibinfo{person}{M. Perrot}, {and}
  \bibinfo{person}{E. Duchesnay}.} \bibinfo{year}{2011}\natexlab{}.
\newblock \showarticletitle{Scikit-learn: Machine Learning in {P}ython}.
\newblock \bibinfo{journal}{\emph{Journal of Machine Learning Research}}
  \bibinfo{volume}{12} (\bibinfo{year}{2011}), \bibinfo{pages}{2825--2830}.
\newblock


\bibitem[Peng et~al\mbox{.}(2023)]%
        {peng2023rwkv}
\bibfield{author}{\bibinfo{person}{Bo Peng}, \bibinfo{person}{Eric Alcaide},
  \bibinfo{person}{Quentin Anthony}, \bibinfo{person}{Alon Albalak},
  \bibinfo{person}{Samuel Arcadinho}, \bibinfo{person}{Huanqi Cao},
  \bibinfo{person}{Xin Cheng}, \bibinfo{person}{Michael Chung},
  \bibinfo{person}{Matteo Grella}, \bibinfo{person}{Kranthi~Kiran GV},
  {et~al\mbox{.}}} \bibinfo{year}{2023}\natexlab{}.
\newblock \showarticletitle{RWKV: Reinventing RNNs for the Transformer Era}.
\newblock \bibinfo{journal}{\emph{arXiv preprint arXiv:2305.13048}}
  (\bibinfo{year}{2023}).
\newblock


\bibitem[Pfaff et~al\mbox{.}(2020)]%
        {ps3}
\bibfield{author}{\bibinfo{person}{Tobias Pfaff}, \bibinfo{person}{Meire
  Fortunato}, \bibinfo{person}{Alvaro Sanchez-Gonzalez}, {and}
  \bibinfo{person}{Peter~W Battaglia}.} \bibinfo{year}{2020}\natexlab{}.
\newblock \showarticletitle{Learning mesh-based simulation with graph
  networks}.
\newblock \bibinfo{journal}{\emph{arXiv preprint arXiv:2010.03409}}
  (\bibinfo{year}{2020}).
\newblock


\bibitem[Poursafaei et~al\mbox{.}(2022)]%
        {evaluation}
\bibfield{author}{\bibinfo{person}{Farimah Poursafaei},
  \bibinfo{person}{Shenyang Huang}, \bibinfo{person}{Kellin Pelrine}, {and}
  \bibinfo{person}{Reihaneh Rabbany}.} \bibinfo{year}{2022}\natexlab{}.
\newblock \showarticletitle{Towards Better Evaluation for Dynamic Link
  Prediction}.
\newblock \bibinfo{journal}{\emph{arXiv preprint arXiv:2207.10128}}
  (\bibinfo{year}{2022}).
\newblock


\bibitem[Radford et~al\mbox{.}(2018)]%
        {gpt}
\bibfield{author}{\bibinfo{person}{Alec Radford}, \bibinfo{person}{Karthik
  Narasimhan}, \bibinfo{person}{Tim Salimans}, \bibinfo{person}{Ilya
  Sutskever}, {et~al\mbox{.}}} \bibinfo{year}{2018}\natexlab{}.
\newblock \showarticletitle{Improving language understanding by generative
  pre-training}.
\newblock  (\bibinfo{year}{2018}).
\newblock


\bibitem[Rae et~al\mbox{.}(2019)]%
        {comptrans}
\bibfield{author}{\bibinfo{person}{Jack~W Rae}, \bibinfo{person}{Anna
  Potapenko}, \bibinfo{person}{Siddhant~M Jayakumar}, {and}
  \bibinfo{person}{Timothy~P Lillicrap}.} \bibinfo{year}{2019}\natexlab{}.
\newblock \showarticletitle{Compressive transformers for long-range sequence
  modelling}.
\newblock \bibinfo{journal}{\emph{arXiv preprint arXiv:1911.05507}}
  (\bibinfo{year}{2019}).
\newblock


\bibitem[Ren et~al\mbox{.}(2023)]%
        {ren2023neural}
\bibfield{author}{\bibinfo{person}{Hongyu Ren}, \bibinfo{person}{Mikhail
  Galkin}, \bibinfo{person}{Michael Cochez}, \bibinfo{person}{Zhaocheng Zhu},
  {and} \bibinfo{person}{Jure Leskovec}.} \bibinfo{year}{2023}\natexlab{}.
\newblock \showarticletitle{Neural graph reasoning: Complex logical query
  answering meets graph databases}.
\newblock \bibinfo{journal}{\emph{arXiv preprint arXiv:2303.14617}}
  (\bibinfo{year}{2023}).
\newblock


\bibitem[Rossi et~al\mbox{.}(2020)]%
        {tgn}
\bibfield{author}{\bibinfo{person}{Emanuele Rossi}, \bibinfo{person}{Ben
  Chamberlain}, \bibinfo{person}{Fabrizio Frasca}, \bibinfo{person}{Davide
  Eynard}, \bibinfo{person}{Federico Monti}, {and} \bibinfo{person}{Michael
  Bronstein}.} \bibinfo{year}{2020}\natexlab{}.
\newblock \showarticletitle{Temporal graph networks for deep learning on
  dynamic graphs}.
\newblock \bibinfo{journal}{\emph{arXiv preprint arXiv:2006.10637}}
  (\bibinfo{year}{2020}).
\newblock


\bibitem[Rossi et~al\mbox{.}(2018a)]%
        {rossiHigherorderNetworkRepresentation2018}
\bibfield{author}{\bibinfo{person}{Ryan~A. Rossi}, \bibinfo{person}{Nesreen~K.
  Ahmed}, {and} \bibinfo{person}{Eunyee Koh}.}
  \bibinfo{year}{2018}\natexlab{a}.
\newblock \showarticletitle{Higher-Order {{Network Representation Learning}}}.
  In \bibinfo{booktitle}{\emph{Companion {{Proceedings}} of the {{The Web
  Conference}} 2018}} \emph{(\bibinfo{series}{{{WWW}} '18})}.
  \bibinfo{publisher}{{International World Wide Web Conferences Steering
  Committee}}, \bibinfo{address}{{Republic and Canton of Geneva, CHE}},
  \bibinfo{pages}{3--4}.
\newblock
\showISBNx{978-1-4503-5640-4}
\urldef\tempurl%
\url{https://doi.org/10.1145/3184558.3186900}
\showDOI{\tempurl}


\bibitem[Rossi et~al\mbox{.}(2018b)]%
        {rossiHONEHigherOrderNetwork2018}
\bibfield{author}{\bibinfo{person}{Ryan~A. Rossi}, \bibinfo{person}{Nesreen~K.
  Ahmed}, \bibinfo{person}{Eunyee Koh}, \bibinfo{person}{Sungchul Kim},
  \bibinfo{person}{Anup Rao}, {and} \bibinfo{person}{Yasin~Abbasi Yadkori}.}
  \bibinfo{year}{2018}\natexlab{b}.
\newblock \showarticletitle{{{HONE}}: {{Higher-Order Network Embeddings}}}.
\newblock \bibinfo{journal}{\emph{arXiv:1801.09303 [cs, stat]}}
  (\bibinfo{date}{May} \bibinfo{year}{2018}).
\newblock
\showeprint[arxiv]{1801.09303}~[cs, stat]


\bibitem[Sakr et~al\mbox{.}(2020)]%
        {sakr2020future}
\bibfield{author}{\bibinfo{person}{Sherif Sakr} {et~al\mbox{.}}}
  \bibinfo{year}{2020}\natexlab{}.
\newblock \showarticletitle{The Future is Big Graphs! A Community View on Graph
  Processing Systems}.
\newblock \bibinfo{journal}{\emph{arXiv preprint arXiv:2012.06171}}
  (\bibinfo{year}{2020}).
\newblock


\bibitem[Sanchez-Gonzalez et~al\mbox{.}(2020)]%
        {sanchez2020learning}
\bibfield{author}{\bibinfo{person}{Alvaro Sanchez-Gonzalez},
  \bibinfo{person}{Jonathan Godwin}, \bibinfo{person}{Tobias Pfaff},
  \bibinfo{person}{Rex Ying}, \bibinfo{person}{Jure Leskovec}, {and}
  \bibinfo{person}{Peter Battaglia}.} \bibinfo{year}{2020}\natexlab{}.
\newblock \showarticletitle{Learning to simulate complex physics with graph
  networks}. In \bibinfo{booktitle}{\emph{ICML}}.
\newblock


\bibitem[Scarselli et~al\mbox{.}(2008)]%
        {scarselli2008graph}
\bibfield{author}{\bibinfo{person}{Franco Scarselli}, \bibinfo{person}{Marco
  Gori}, \bibinfo{person}{Ah~Chung Tsoi}, \bibinfo{person}{Markus
  Hagenbuchner}, {and} \bibinfo{person}{Gabriele Monfardini}.}
  \bibinfo{year}{2008}\natexlab{}.
\newblock \showarticletitle{The graph neural network model}.
\newblock \bibinfo{journal}{\emph{IEEE transactions on neural networks}}
  \bibinfo{volume}{20}, \bibinfo{number}{1} (\bibinfo{year}{2008}),
  \bibinfo{pages}{61--80}.
\newblock


\bibitem[Shazeer(2020)]%
        {geglu}
\bibfield{author}{\bibinfo{person}{Noam Shazeer}.}
  \bibinfo{year}{2020}\natexlab{}.
\newblock \showarticletitle{Glu variants improve transformer}.
\newblock \bibinfo{journal}{\emph{arXiv preprint arXiv:2002.05202}}
  (\bibinfo{year}{2020}).
\newblock


\bibitem[Sizemore and Bassett(2018)]%
        {sizemore2018dynamic}
\bibfield{author}{\bibinfo{person}{Ann~E Sizemore} {and}
  \bibinfo{person}{Danielle~S Bassett}.} \bibinfo{year}{2018}\natexlab{}.
\newblock \showarticletitle{Dynamic graph metrics: Tutorial, toolbox, and
  tale}.
\newblock \bibinfo{journal}{\emph{NeuroImage}}  \bibinfo{volume}{180}
  (\bibinfo{year}{2018}), \bibinfo{pages}{417--427}.
\newblock


\bibitem[Skarding et~al\mbox{.}(2021)]%
        {skarding2021foundations}
\bibfield{author}{\bibinfo{person}{Joakim Skarding}, \bibinfo{person}{Bogdan
  Gabrys}, {and} \bibinfo{person}{Katarzyna Musial}.}
  \bibinfo{year}{2021}\natexlab{}.
\newblock \showarticletitle{Foundations and modeling of dynamic networks using
  dynamic graph neural networks: A survey}.
\newblock \bibinfo{journal}{\emph{IEEE Access}}  \bibinfo{volume}{9}
  (\bibinfo{year}{2021}), \bibinfo{pages}{79143--79168}.
\newblock


\bibitem[Song et~al\mbox{.}(2019)]%
        {socNet2}
\bibfield{author}{\bibinfo{person}{Weiping Song}, \bibinfo{person}{Zhiping
  Xiao}, \bibinfo{person}{Yifan Wang}, \bibinfo{person}{Laurent Charlin},
  \bibinfo{person}{Ming Zhang}, {and} \bibinfo{person}{Jian Tang}.}
  \bibinfo{year}{2019}\natexlab{}.
\newblock \showarticletitle{Session-based social recommendation via dynamic
  graph attention networks}. In \bibinfo{booktitle}{\emph{Proceedings of the
  Twelfth ACM international conference on web search and data mining}}.
  \bibinfo{pages}{555--563}.
\newblock


\bibitem[Sukhbaatar et~al\mbox{.}(2016)]%
        {sukhbaatar2016learning}
\bibfield{author}{\bibinfo{person}{Sainbayar Sukhbaatar}, \bibinfo{person}{Rob
  Fergus}, {et~al\mbox{.}}} \bibinfo{year}{2016}\natexlab{}.
\newblock \showarticletitle{Learning multiagent communication with
  backpropagation}.
\newblock \bibinfo{journal}{\emph{NeurIPS}} (\bibinfo{year}{2016}).
\newblock


\bibitem[Sukhbaatar et~al\mbox{.}(2015)]%
        {mn2}
\bibfield{author}{\bibinfo{person}{Sainbayar Sukhbaatar},
  \bibinfo{person}{arthur szlam}, \bibinfo{person}{Jason Weston}, {and}
  \bibinfo{person}{Rob Fergus}.} \bibinfo{year}{2015}\natexlab{}.
\newblock \showarticletitle{End-To-End Memory Networks}. In
  \bibinfo{booktitle}{\emph{Advances in Neural Information Processing
  Systems}}, \bibfield{editor}{\bibinfo{person}{C.~Cortes},
  \bibinfo{person}{N.~Lawrence}, \bibinfo{person}{D.~Lee},
  \bibinfo{person}{M.~Sugiyama}, {and} \bibinfo{person}{R.~Garnett}} (Eds.),
  Vol.~\bibinfo{volume}{28}. \bibinfo{publisher}{Curran Associates, Inc.}
\newblock
\urldef\tempurl%
\url{https://proceedings.neurips.cc/paper_files/paper/2015/file/8fb21ee7a2207526da55a679f0332de2-Paper.pdf}
\showURL{%
\tempurl}


\bibitem[Sun et~al\mbox{.}(2022)]%
        {xpos}
\bibfield{author}{\bibinfo{person}{Yutao Sun}, \bibinfo{person}{Li Dong},
  \bibinfo{person}{Barun Patra}, \bibinfo{person}{Shuming Ma},
  \bibinfo{person}{Shaohan Huang}, \bibinfo{person}{Alon Benhaim},
  \bibinfo{person}{Vishrav Chaudhary}, \bibinfo{person}{Xia Song}, {and}
  \bibinfo{person}{Furu Wei}.} \bibinfo{year}{2022}\natexlab{}.
\newblock \showarticletitle{A length-extrapolatable transformer}.
\newblock \bibinfo{journal}{\emph{arXiv preprint arXiv:2212.10554}}
  (\bibinfo{year}{2022}).
\newblock


\bibitem[Sutskever et~al\mbox{.}(2014)]%
        {rnn1}
\bibfield{author}{\bibinfo{person}{Ilya Sutskever}, \bibinfo{person}{Oriol
  Vinyals}, {and} \bibinfo{person}{Quoc~V Le}.}
  \bibinfo{year}{2014}\natexlab{}.
\newblock \showarticletitle{Sequence to Sequence Learning with Neural
  Networks}. In \bibinfo{booktitle}{\emph{Advances in Neural Information
  Processing Systems}}, \bibfield{editor}{\bibinfo{person}{Z.~Ghahramani},
  \bibinfo{person}{M.~Welling}, \bibinfo{person}{C.~Cortes},
  \bibinfo{person}{N.~Lawrence}, {and} \bibinfo{person}{K.Q. Weinberger}}
  (Eds.), Vol.~\bibinfo{volume}{27}. \bibinfo{publisher}{Curran Associates,
  Inc.}
\newblock


\bibitem[Thekumparampil et~al\mbox{.}(2018)]%
        {thekumparampil2018attention}
\bibfield{author}{\bibinfo{person}{Kiran~K Thekumparampil},
  \bibinfo{person}{Chong Wang}, \bibinfo{person}{Sewoong Oh}, {and}
  \bibinfo{person}{Li-Jia Li}.} \bibinfo{year}{2018}\natexlab{}.
\newblock \showarticletitle{Attention-based graph neural network for
  semi-supervised learning}.
\newblock \bibinfo{journal}{\emph{arXiv preprint arXiv:1803.03735}}
  (\bibinfo{year}{2018}).
\newblock


\bibitem[Tolstikhin et~al\mbox{.}(2021)]%
        {mlp-mixer}
\bibfield{author}{\bibinfo{person}{Ilya~O Tolstikhin}, \bibinfo{person}{Neil
  Houlsby}, \bibinfo{person}{Alexander Kolesnikov}, \bibinfo{person}{Lucas
  Beyer}, \bibinfo{person}{Xiaohua Zhai}, \bibinfo{person}{Thomas Unterthiner},
  \bibinfo{person}{Jessica Yung}, \bibinfo{person}{Andreas Steiner},
  \bibinfo{person}{Daniel Keysers}, \bibinfo{person}{Jakob Uszkoreit},
  {et~al\mbox{.}}} \bibinfo{year}{2021}\natexlab{}.
\newblock \showarticletitle{Mlp-mixer: An all-mlp architecture for vision}.
\newblock \bibinfo{journal}{\emph{Advances in neural information processing
  systems}}  \bibinfo{volume}{34} (\bibinfo{year}{2021}),
  \bibinfo{pages}{24261--24272}.
\newblock


\bibitem[Torres et~al\mbox{.}(2021)]%
        {torres2021and}
\bibfield{author}{\bibinfo{person}{Leo Torres}, \bibinfo{person}{Ann~S
  Blevins}, \bibinfo{person}{Danielle Bassett}, {and} \bibinfo{person}{Tina
  Eliassi-Rad}.} \bibinfo{year}{2021}\natexlab{}.
\newblock \showarticletitle{The why, how, and when of representations for
  complex systems}.
\newblock \bibinfo{journal}{\emph{SIAM Rev.}} \bibinfo{volume}{63},
  \bibinfo{number}{3} (\bibinfo{year}{2021}), \bibinfo{pages}{435--485}.
\newblock


\bibitem[Tripathy et~al\mbox{.}(2020)]%
        {tripathy2020reducing}
\bibfield{author}{\bibinfo{person}{Alok Tripathy}, \bibinfo{person}{Katherine
  Yelick}, {and} \bibinfo{person}{Ayd{\i}n Bulu{\c{c}}}.}
  \bibinfo{year}{2020}\natexlab{}.
\newblock \showarticletitle{Reducing communication in graph neural network
  training}. In \bibinfo{booktitle}{\emph{ACM/IEEE Supercomputing}}.
\newblock


\bibitem[Trivedi et~al\mbox{.}(2019)]%
        {dyrep}
\bibfield{author}{\bibinfo{person}{Rakshit Trivedi}, \bibinfo{person}{Mehrdad
  Farajtabar}, \bibinfo{person}{Prasenjeet Biswal}, {and}
  \bibinfo{person}{Hongyuan Zha}.} \bibinfo{year}{2019}\natexlab{}.
\newblock \showarticletitle{Dyrep: Learning representations over dynamic
  graphs}. In \bibinfo{booktitle}{\emph{International conference on learning
  representations}}.
\newblock


\bibitem[Vaswani et~al\mbox{.}(2017)]%
        {vaswani2017attention}
\bibfield{author}{\bibinfo{person}{Ashish Vaswani}, \bibinfo{person}{Noam
  Shazeer}, \bibinfo{person}{Niki Parmar}, \bibinfo{person}{Jakob Uszkoreit},
  \bibinfo{person}{Llion Jones}, \bibinfo{person}{Aidan~N Gomez},
  \bibinfo{person}{{\L}ukasz Kaiser}, {and} \bibinfo{person}{Illia
  Polosukhin}.} \bibinfo{year}{2017}\natexlab{}.
\newblock \showarticletitle{Attention is all you need}. In
  \bibinfo{booktitle}{\emph{NeurIPS}}.
\newblock


\bibitem[Veli{\v{c}}kovi{\'c} et~al\mbox{.}(2017)]%
        {velivckovic2017graph}
\bibfield{author}{\bibinfo{person}{Petar Veli{\v{c}}kovi{\'c}},
  \bibinfo{person}{Guillem Cucurull}, \bibinfo{person}{Arantxa Casanova},
  \bibinfo{person}{Adriana Romero}, \bibinfo{person}{Pietro Lio}, {and}
  \bibinfo{person}{Yoshua Bengio}.} \bibinfo{year}{2017}\natexlab{}.
\newblock \showarticletitle{Graph attention networks}.
\newblock \bibinfo{journal}{\emph{arXiv preprint arXiv:1710.10903}}
  (\bibinfo{year}{2017}).
\newblock


\bibitem[Waleffe et~al\mbox{.}(2022)]%
        {waleffe2022marius++}
\bibfield{author}{\bibinfo{person}{Roger Waleffe}, \bibinfo{person}{Jason
  Mohoney}, \bibinfo{person}{Theodoros Rekatsinas}, {and}
  \bibinfo{person}{Shivaram Venkataraman}.} \bibinfo{year}{2022}\natexlab{}.
\newblock \showarticletitle{Marius++: Large-Scale Training of Graph Neural
  Networks on a Single Machine}.
\newblock \bibinfo{journal}{\emph{arXiv preprint arXiv:2202.02365}}
  (\bibinfo{year}{2022}).
\newblock


\bibitem[Wan et~al\mbox{.}(2022a)]%
        {wan2022bns}
\bibfield{author}{\bibinfo{person}{Cheng Wan}, \bibinfo{person}{Youjie Li},
  \bibinfo{person}{Ang Li}, \bibinfo{person}{Nam~Sung Kim}, {and}
  \bibinfo{person}{Yingyan Lin}.} \bibinfo{year}{2022}\natexlab{a}.
\newblock \showarticletitle{BNS-GCN: Efficient Full-Graph Training of Graph
  Convolutional Networks with Partition-Parallelism and Random Boundary Node
  Sampling Sampling}.
\newblock \bibinfo{journal}{\emph{MLSys}} (\bibinfo{year}{2022}).
\newblock


\bibitem[Wan et~al\mbox{.}(2022b)]%
        {wan2022pipegcn}
\bibfield{author}{\bibinfo{person}{Cheng Wan}, \bibinfo{person}{Youjie Li},
  \bibinfo{person}{Cameron~R Wolfe}, \bibinfo{person}{Anastasios Kyrillidis},
  \bibinfo{person}{Nam~Sung Kim}, {and} \bibinfo{person}{Yingyan Lin}.}
  \bibinfo{year}{2022}\natexlab{b}.
\newblock \showarticletitle{PipeGCN: Efficient full-graph training of graph
  convolutional networks with pipelined feature communication}.
\newblock \bibinfo{journal}{\emph{arXiv preprint arXiv:2203.10428}}
  (\bibinfo{year}{2022}).
\newblock


\bibitem[Wang et~al\mbox{.}(2021a)]%
        {tcl}
\bibfield{author}{\bibinfo{person}{Lu Wang}, \bibinfo{person}{Xiaofu Chang},
  \bibinfo{person}{Shuang Li}, \bibinfo{person}{Yunfei Chu},
  \bibinfo{person}{Hui Li}, \bibinfo{person}{Wei Zhang},
  \bibinfo{person}{Xiaofeng He}, \bibinfo{person}{Le Song},
  \bibinfo{person}{Jingren Zhou}, {and} \bibinfo{person}{Hongxia Yang}.}
  \bibinfo{year}{2021}\natexlab{a}.
\newblock \showarticletitle{Tcl: Transformer-based dynamic graph modelling via
  contrastive learning}.
\newblock \bibinfo{journal}{\emph{arXiv preprint arXiv:2105.07944}}
  (\bibinfo{year}{2021}).
\newblock


\bibitem[Wang et~al\mbox{.}(2019b)]%
        {wang2019deep}
\bibfield{author}{\bibinfo{person}{Minjie Wang}, \bibinfo{person}{Da Zheng},
  \bibinfo{person}{Zihao Ye}, \bibinfo{person}{Quan Gan},
  \bibinfo{person}{Mufei Li}, \bibinfo{person}{Xiang Song},
  \bibinfo{person}{Jinjing Zhou}, \bibinfo{person}{Chao Ma},
  \bibinfo{person}{Lingfan Yu}, \bibinfo{person}{Yu Gai}, {et~al\mbox{.}}}
  \bibinfo{year}{2019}\natexlab{b}.
\newblock \showarticletitle{Deep graph library: A graph-centric,
  highly-performant package for graph neural networks}.
\newblock \bibinfo{journal}{\emph{arXiv:1909.01315}} (\bibinfo{year}{2019}).
\newblock


\bibitem[Wang et~al\mbox{.}(2021d)]%
        {con9}
\bibfield{author}{\bibinfo{person}{Xuhong Wang}, \bibinfo{person}{Ding Lyu},
  \bibinfo{person}{Mengjian Li}, \bibinfo{person}{Yang Xia},
  \bibinfo{person}{Qi Yang}, \bibinfo{person}{Xinwen Wang},
  \bibinfo{person}{Xinguang Wang}, \bibinfo{person}{Ping Cui},
  \bibinfo{person}{Yupu Yang}, \bibinfo{person}{Bowen Sun}, {et~al\mbox{.}}}
  \bibinfo{year}{2021}\natexlab{d}.
\newblock \showarticletitle{Apan: Asynchronous propagation attention network
  for real-time temporal graph embedding}. In
  \bibinfo{booktitle}{\emph{Proceedings of the 2021 international conference on
  management of data}}. \bibinfo{pages}{2628--2638}.
\newblock


\bibitem[Wang et~al\mbox{.}(2021c)]%
        {wang2020gnnadvisor}
\bibfield{author}{\bibinfo{person}{Yuke Wang} {et~al\mbox{.}}}
  \bibinfo{year}{2021}\natexlab{c}.
\newblock \showarticletitle{GNNAdvisor: An Efficient Runtime System for GNN
  Acceleration on GPUs}. In \bibinfo{booktitle}{\emph{OSDI}}.
\newblock


\bibitem[Wang et~al\mbox{.}(2021b)]%
        {cawn}
\bibfield{author}{\bibinfo{person}{Yanbang Wang}, \bibinfo{person}{Yen-Yu
  Chang}, \bibinfo{person}{Yunyu Liu}, \bibinfo{person}{Jure Leskovec}, {and}
  \bibinfo{person}{Pan Li}.} \bibinfo{year}{2021}\natexlab{b}.
\newblock \showarticletitle{Inductive representation learning in temporal
  networks via causal anonymous walks}.
\newblock \bibinfo{journal}{\emph{arXiv preprint arXiv:2101.05974}}
  (\bibinfo{year}{2021}).
\newblock


\bibitem[Wang et~al\mbox{.}(2019a)]%
        {wang2019dynamic}
\bibfield{author}{\bibinfo{person}{Yue Wang}, \bibinfo{person}{Yongbin Sun},
  \bibinfo{person}{Ziwei Liu}, \bibinfo{person}{Sanjay~E Sarma},
  \bibinfo{person}{Michael~M Bronstein}, {and} \bibinfo{person}{Justin~M
  Solomon}.} \bibinfo{year}{2019}\natexlab{a}.
\newblock \showarticletitle{Dynamic graph cnn for learning on point clouds}.
\newblock \bibinfo{journal}{\emph{Acm Transactions On Graphics (tog)}}
  \bibinfo{volume}{38}, \bibinfo{number}{5} (\bibinfo{year}{2019}),
  \bibinfo{pages}{1--12}.
\newblock


\bibitem[Wu et~al\mbox{.}(2021b)]%
        {fastformer}
\bibfield{author}{\bibinfo{person}{Chuhan Wu}, \bibinfo{person}{Fangzhao Wu},
  \bibinfo{person}{Tao Qi}, \bibinfo{person}{Yongfeng Huang}, {and}
  \bibinfo{person}{Xing Xie}.} \bibinfo{year}{2021}\natexlab{b}.
\newblock \showarticletitle{Fastformer: Additive attention can be all you
  need}.
\newblock \bibinfo{journal}{\emph{arXiv preprint arXiv:2108.09084}}
  (\bibinfo{year}{2021}).
\newblock


\bibitem[Wu et~al\mbox{.}(2019b)]%
        {wu2019simplifying}
\bibfield{author}{\bibinfo{person}{Felix Wu}, \bibinfo{person}{Amauri Souza},
  \bibinfo{person}{Tianyi Zhang}, \bibinfo{person}{Christopher Fifty},
  \bibinfo{person}{Tao Yu}, {and} \bibinfo{person}{Kilian Weinberger}.}
  \bibinfo{year}{2019}\natexlab{b}.
\newblock \showarticletitle{Simplifying graph convolutional networks}. In
  \bibinfo{booktitle}{\emph{International conference on machine learning}}.
  PMLR, \bibinfo{pages}{6861--6871}.
\newblock


\bibitem[Wu et~al\mbox{.}(2021a)]%
        {wu2021seastar}
\bibfield{author}{\bibinfo{person}{Yidi Wu}, \bibinfo{person}{Kaihao Ma},
  \bibinfo{person}{Zhenkun Cai}, \bibinfo{person}{Tatiana Jin},
  \bibinfo{person}{Boyang Li}, \bibinfo{person}{Chenguang Zheng},
  \bibinfo{person}{James Cheng}, {and} \bibinfo{person}{Fan Yu}.}
  \bibinfo{year}{2021}\natexlab{a}.
\newblock \showarticletitle{Seastar: vertex-centric programming for graph
  neural networks}. In \bibinfo{booktitle}{\emph{EuroSys}}.
\newblock


\bibitem[Wu et~al\mbox{.}(2020)]%
        {wu2020comprehensive}
\bibfield{author}{\bibinfo{person}{Zonghan Wu} {et~al\mbox{.}}}
  \bibinfo{year}{2020}\natexlab{}.
\newblock \showarticletitle{A comprehensive survey on graph neural networks}.
\newblock \bibinfo{journal}{\emph{IEEE Transactions on Neural Networks and
  Learning Systems}} (\bibinfo{year}{2020}).
\newblock


\bibitem[Wu et~al\mbox{.}(2019a)]%
        {tn2}
\bibfield{author}{\bibinfo{person}{Zonghan Wu}, \bibinfo{person}{Shirui Pan},
  \bibinfo{person}{Guodong Long}, \bibinfo{person}{Jing Jiang}, {and}
  \bibinfo{person}{Chengqi Zhang}.} \bibinfo{year}{2019}\natexlab{a}.
\newblock \showarticletitle{Graph wavenet for deep spatial-temporal graph
  modeling}.
\newblock \bibinfo{journal}{\emph{arXiv preprint arXiv:1906.00121}}
  (\bibinfo{year}{2019}).
\newblock


\bibitem[Xu et~al\mbox{.}(2020)]%
        {tgat}
\bibfield{author}{\bibinfo{person}{Da Xu}, \bibinfo{person}{Chuanwei Ruan},
  \bibinfo{person}{Evren Korpeoglu}, \bibinfo{person}{Sushant Kumar}, {and}
  \bibinfo{person}{Kannan Achan}.} \bibinfo{year}{2020}\natexlab{}.
\newblock \showarticletitle{Inductive representation learning on temporal
  graphs}.
\newblock \bibinfo{journal}{\emph{arXiv preprint arXiv:2002.07962}}
  (\bibinfo{year}{2020}).
\newblock


\bibitem[Xu et~al\mbox{.}(2018)]%
        {xu2018powerful}
\bibfield{author}{\bibinfo{person}{Keyulu Xu}, \bibinfo{person}{Weihua Hu},
  \bibinfo{person}{Jure Leskovec}, {and} \bibinfo{person}{Stefanie Jegelka}.}
  \bibinfo{year}{2018}\natexlab{}.
\newblock \showarticletitle{How powerful are graph neural networks?}
\newblock \bibinfo{journal}{\emph{arXiv preprint arXiv:1810.00826}}
  (\bibinfo{year}{2018}).
\newblock


\bibitem[Xue et~al\mbox{.}(2022)]%
        {xue2022dynamic}
\bibfield{author}{\bibinfo{person}{Guotong Xue}, \bibinfo{person}{Ming Zhong},
  \bibinfo{person}{Jianxin Li}, \bibinfo{person}{Jia Chen},
  \bibinfo{person}{Chengshuai Zhai}, {and} \bibinfo{person}{Ruochen Kong}.}
  \bibinfo{year}{2022}\natexlab{}.
\newblock \showarticletitle{Dynamic network embedding survey}.
\newblock \bibinfo{journal}{\emph{Neurocomputing}}  \bibinfo{volume}{472}
  (\bibinfo{year}{2022}), \bibinfo{pages}{212--223}.
\newblock


\bibitem[Yan et~al\mbox{.}(2020)]%
        {yan2020hygcn}
\bibfield{author}{\bibinfo{person}{Mingyu Yan}, \bibinfo{person}{Lei Deng},
  \bibinfo{person}{Xing Hu}, \bibinfo{person}{Ling Liang},
  \bibinfo{person}{Yujing Feng}, \bibinfo{person}{Xiaochun Ye},
  \bibinfo{person}{Zhimin Zhang}, \bibinfo{person}{Dongrui Fan}, {and}
  \bibinfo{person}{Yuan Xie}.} \bibinfo{year}{2020}\natexlab{}.
\newblock \showarticletitle{Hygcn: A gcn accelerator with hybrid architecture}.
  In \bibinfo{booktitle}{\emph{IEEE HPCA}}. IEEE, \bibinfo{pages}{15--29}.
\newblock


\bibitem[Yu et~al\mbox{.}(2017)]%
        {tn1}
\bibfield{author}{\bibinfo{person}{Bing Yu}, \bibinfo{person}{Haoteng Yin},
  {and} \bibinfo{person}{Zhanxing Zhu}.} \bibinfo{year}{2017}\natexlab{}.
\newblock \showarticletitle{Spatio-temporal graph convolutional networks: A
  deep learning framework for traffic forecasting}.
\newblock \bibinfo{journal}{\emph{arXiv preprint arXiv:1709.04875}}
  (\bibinfo{year}{2017}).
\newblock


\bibitem[Yu et~al\mbox{.}(2021)]%
        {tn5}
\bibfield{author}{\bibinfo{person}{Le Yu}, \bibinfo{person}{Bowen Du},
  \bibinfo{person}{Xiao Hu}, \bibinfo{person}{Leilei Sun},
  \bibinfo{person}{Liangzhe Han}, {and} \bibinfo{person}{Weifeng Lv}.}
  \bibinfo{year}{2021}\natexlab{}.
\newblock \showarticletitle{Deep spatio-temporal graph convolutional network
  for traffic accident prediction}.
\newblock \bibinfo{journal}{\emph{Neurocomputing}}  \bibinfo{volume}{423}
  (\bibinfo{year}{2021}), \bibinfo{pages}{135--147}.
\newblock


\bibitem[Yu et~al\mbox{.}(2022a)]%
        {ui5}
\bibfield{author}{\bibinfo{person}{Le Yu}, \bibinfo{person}{Zihang Liu},
  \bibinfo{person}{Tongyu Zhu}, \bibinfo{person}{Leilei Sun},
  \bibinfo{person}{Bowen Du}, {and} \bibinfo{person}{Weifeng Lv}.}
  \bibinfo{year}{2022}\natexlab{a}.
\newblock \showarticletitle{Modelling Evolutionary and Stationary User
  Preferences for Temporal Sets Prediction}.
\newblock \bibinfo{journal}{\emph{arXiv preprint arXiv:2204.05490}}
  (\bibinfo{year}{2022}).
\newblock


\bibitem[Yu et~al\mbox{.}(2023)]%
        {dygformer}
\bibfield{author}{\bibinfo{person}{Le Yu}, \bibinfo{person}{Leilei Sun},
  \bibinfo{person}{Bowen Du}, {and} \bibinfo{person}{Weifeng Lv}.}
  \bibinfo{year}{2023}\natexlab{}.
\newblock \bibinfo{title}{Towards Better Dynamic Graph Learning: New
  Architecture and Unified Library}.
\newblock
\newblock
\showeprint[arxiv]{2303.13047}~[cs.LG]


\bibitem[Yu et~al\mbox{.}(2022b)]%
        {ui3}
\bibfield{author}{\bibinfo{person}{Le Yu}, \bibinfo{person}{Guanghui Wu},
  \bibinfo{person}{Leilei Sun}, \bibinfo{person}{Bowen Du}, {and}
  \bibinfo{person}{Weifeng Lv}.} \bibinfo{year}{2022}\natexlab{b}.
\newblock \showarticletitle{Element-guided Temporal Graph Representation
  Learning for Temporal Sets Prediction}. In
  \bibinfo{booktitle}{\emph{Proceedings of the ACM Web Conference 2022}}.
  \bibinfo{pages}{1902--1913}.
\newblock


\bibitem[Zhang et~al\mbox{.}(2020b)]%
        {zhang2020agl}
\bibfield{author}{\bibinfo{person}{Dalong Zhang} {et~al\mbox{.}}}
  \bibinfo{year}{2020}\natexlab{b}.
\newblock \showarticletitle{Agl: a scalable system for industrial-purpose graph
  machine learning}.
\newblock \bibinfo{journal}{\emph{arXiv preprint arXiv:2003.02454}}
  (\bibinfo{year}{2020}).
\newblock


\bibitem[Zhang and Chen(2018)]%
        {zhang2018link}
\bibfield{author}{\bibinfo{person}{Muhan Zhang} {and} \bibinfo{person}{Yixin
  Chen}.} \bibinfo{year}{2018}\natexlab{}.
\newblock \showarticletitle{Link prediction based on graph neural networks}.
\newblock \bibinfo{journal}{\emph{arXiv preprint arXiv:1802.09691}}
  (\bibinfo{year}{2018}).
\newblock


\bibitem[Zhang et~al\mbox{.}(2022a)]%
        {ui4}
\bibfield{author}{\bibinfo{person}{Mengqi Zhang}, \bibinfo{person}{Shu Wu},
  \bibinfo{person}{Xueli Yu}, \bibinfo{person}{Qiang Liu}, {and}
  \bibinfo{person}{Liang Wang}.} \bibinfo{year}{2022}\natexlab{a}.
\newblock \showarticletitle{Dynamic graph neural networks for sequential
  recommendation}.
\newblock \bibinfo{journal}{\emph{IEEE Transactions on Knowledge and Data
  Engineering}} \bibinfo{volume}{35}, \bibinfo{number}{5}
  (\bibinfo{year}{2022}), \bibinfo{pages}{4741--4753}.
\newblock


\bibitem[Zhang et~al\mbox{.}(2020a)]%
        {zhang2020deep}
\bibfield{author}{\bibinfo{person}{Ziwei Zhang}, \bibinfo{person}{Peng Cui},
  {and} \bibinfo{person}{Wenwu Zhu}.} \bibinfo{year}{2020}\natexlab{a}.
\newblock \showarticletitle{Deep learning on graphs: A survey}.
\newblock \bibinfo{journal}{\emph{IEEE Transactions on Knowledge and Data
  Engineering}} (\bibinfo{year}{2020}).
\newblock


\bibitem[Zhang et~al\mbox{.}(2022b)]%
        {zhang2022nested}
\bibfield{author}{\bibinfo{person}{Zizhao Zhang}, \bibinfo{person}{Han Zhang},
  \bibinfo{person}{Long Zhao}, \bibinfo{person}{Ting Chen},
  \bibinfo{person}{Sercan~{\"O} Arik}, {and} \bibinfo{person}{Tomas Pfister}.}
  \bibinfo{year}{2022}\natexlab{b}.
\newblock \showarticletitle{Nested hierarchical transformer: Towards accurate,
  data-efficient and interpretable visual understanding}. In
  \bibinfo{booktitle}{\emph{Proceedings of the AAAI Conference on Artificial
  Intelligence}}, Vol.~\bibinfo{volume}{36}. \bibinfo{pages}{3417--3425}.
\newblock


\bibitem[Zheng et~al\mbox{.}(2021)]%
        {zheng2021distributed}
\bibfield{author}{\bibinfo{person}{Da Zheng}, \bibinfo{person}{Xiang Song},
  \bibinfo{person}{Chengru Yang}, \bibinfo{person}{Dominique LaSalle},
  \bibinfo{person}{Qidong Su}, \bibinfo{person}{Minjie Wang},
  \bibinfo{person}{Chao Ma}, {and} \bibinfo{person}{George Karypis}.}
  \bibinfo{year}{2021}\natexlab{}.
\newblock \showarticletitle{Distributed Hybrid CPU and GPU training for Graph
  Neural Networks on Billion-Scale Graphs}.
\newblock \bibinfo{journal}{\emph{arXiv:2112.15345}} (\bibinfo{year}{2021}).
\newblock


\bibitem[Zhou et~al\mbox{.}(2020)]%
        {zhou2020graph}
\bibfield{author}{\bibinfo{person}{Jie Zhou} {et~al\mbox{.}}}
  \bibinfo{year}{2020}\natexlab{}.
\newblock \showarticletitle{Graph neural networks: A review of methods and
  applications}.
\newblock \bibinfo{journal}{\emph{AI Open}}  \bibinfo{volume}{1}
  (\bibinfo{year}{2020}), \bibinfo{pages}{57--81}.
\newblock


\bibitem[Zhu et~al\mbox{.}(2019)]%
        {zhu2019aligraph}
\bibfield{author}{\bibinfo{person}{Rong Zhu} {et~al\mbox{.}}}
  \bibinfo{year}{2019}\natexlab{}.
\newblock \showarticletitle{Aligraph: A comprehensive graph neural network
  platform}.
\newblock \bibinfo{journal}{\emph{arXiv preprint arXiv:1902.08730}}
  (\bibinfo{year}{2019}).
\newblock


\end{thebibliography}
}

%\newpage

%\pagenumbering{gobble} 

\appendix

\section*{Appendix}

\section{{Model Design: Additional Details}}

\marginpar{\large\vspace{-1em}{\textbf{Audi}}\\{\textbf{(2.2)}}}

\hl{We provide more details on the model design.}

\subsection{{Vanilla Transformer}}

\hl{We detail some parts of the vanilla Transformer that are used when constructing the Block-Recurrent Transformer.}

\subsubsection{{Positional Encoding}}

\hl{Before the sequence arrives at the encoder, it requires positional encoding, such that the Transformer has a notion of the order of the inputs $x_i$ in the input sequence. The original paper suggests the following transformation of the inputs:}

\begin{equation}
    Z_{i, j} = X_{i, j} + 
    \begin{cases*}
     \sin \left( i / 1000^{j/d} \right) & if $j$ even, \\
     \cos \left( i / 1000^{\left( j - 1 \right)/d} \right) & otherwise. \\
    \end{cases*}
\end{equation}

\hl{The matrix $Z$ is then fed into the multi-head self-attention module of the encoder.}

\subsubsection{{Multi-Head Self-Attention}}
\label{sec:mh}

\hl{Here, the matrix $Z$ is first projected onto three different matrices $Q = ZW_Q \in \mathbb{R}^{n \times d_k}$, $K = ZW_K \in \mathbb{R}^{n \times d_k}$ and $V = ZW_V \in \mathbb{R}^{n \times d_v}$, where $W_Q$, $W_K$ and $W_V$ are parameter matrices. The matrix of outputs is then calculated as follows:}

\begin{equation}
    \text{Attention} \left( Q, K, V \right) = \text{softmax} \left( \frac{QK^T}{\sqrt{d}} \right) V
\end{equation}

\hl{In order to capture information from different representation subspaces we can perform Multi-Head Attention. In this case, the matrix of outputs is calculated as follows:}

\begin{equation}
    \text{MH} \left( Q, K, V \right) = \text{Concat} \left( \text{head}_1, \ldots, \text{head}_h \right) W_O
\end{equation}

\hl{where $W_O \in \mathbb{R}^{hd_v \times d}$, and}

\begin{equation}
    \text{head}_i = \text{Attention} \left( QW_Q^i, KW_K^i, VW_V^i \right)
\end{equation}

\hl{where $W_Q^i \in \mathbb{R}^{d \times d_k}$, $W_K^i \in \mathbb{R}^{d \times d_k}$, $W_V^i \in \mathbb{R}^{d \times d_v}$ are parameter matrices.}

\hl{The output of this module is then added to $Z$ over a residual connection \mbox{\cite{rc}} and normalised using Layer Normalisation \mbox{\cite{ln}}. It is then fed into an MLP.}

\subsubsection{{Feed-Forward Network}}

\hl{The MLP consists of a single ReLu-activated hidden layer. It is applied to each element of the sequence separately. The output is then, as before, added to the output of the previous module over a residual connection and normalised using Layer Normalisation.}

\hl{The output of the encoder layer is then fed back into another encoder layer. The number of layers in the encoder depends on the implementation. The original paper suggests a total of 6 layers. Afterwards, the output is fed into the decoder, which computes the final output probabilities.}

\subsection{{Block-Recurrent Transformer}}

\subsubsection{{Model Overview}}
\label{sec:brt}

\hl{The first block within each segment must be able to attend to previous segments. As such, following an idea from Transformer-XL \mbox{\cite{transxl}}, the model stores the keys and values of the computed state vectors of the previous segment in cache. }

\if 0
\begin{figure}[ht]
 \centering
 \includegraphics[width=\textwidth]{brt}
\caption{Architecture of the vertical (left) and horizontal (right) directions of the recurrent cells of the Block-Recurrent Transformer model. (Figure taken from \cite{brt})}
\label{fig:brt}
\end{figure}
\fi

\hl{The Block-Recurrent Transformer is composed of various modules called recurrent cells. There are two types of recurrent cells, vertical cells, which calculate next state vectors, and horizontal cells, which calculate output token embeddings. These can be stacked in any desired way. The blocks are fed one by one to the stack of recurrent cells, and the outputs of the last cell are then concatenated to form the output of the model.}

% A short description of each type of recurrent cell follows. A pictorial representation of the cells' architectures can be found in Figure \ref{fig:brt}.

\paragraph{{Vertical cell:}} \hl{Just like in the vanilla Transformer, in the vertical recurrent cell, the input token embeddings undergo self-attention. However, unlike the vanilla Transformer, this type of cell also employs cross-attention on the input embeddings and the current state vectors, i.e., it computes attention scores using a query matrix ($Q$) based on the input embeddings and key and value matrices ($K$, $V$) based on the current state vectors. The results from both attention modules are then projected to some feature space and added to the input token embeddings over a residual connection. The resulting values are then fed into an MLP and added to its output to form the output token embeddings.}

\paragraph{{Horizontal cell:}} \hl{In the horizontal cell the roles of the current state vectors and the input token embeddings are inverted. While the current state vectors undergo self-attention and contribute to the cross-attention module with a query matrix ($Q$), the input token embeddings only contribute to the cross-attention module with the key and value matrices ($K$, $V$). The results of the attention modules are projected onto some feature space. The resulting values $h_t$ are then fed into a group of gates, one per current state vector $c_t$. Note, that $t$ refers to the index of the current block within the sequence. Each gate realizes the following computations:}

\begin{equation}
    z_t = W_zh_t + b_z
\end{equation}
\begin{equation}
    g = \sigma (b_g)
\end{equation}
\begin{equation}
    c_{t + 1} = c_t \odot g + z_t \odot (1 - g)
\end{equation}

\hl{where $W_z$, $b_z$ and $b_g$ are trainable parameters, $\sigma$ is the sigmoid function and $\odot$ is the element-wise multiplication. The outputs of the gates $c_{t+1}$ are then concatenated to form the next state vectors and the output of this cell.}

\if 0

Note, that the architecture description of the horizontal cell above differs slightly from the one depicted in Figure \ref{fig:brt}. The figure presents an additional MLP as well as an additional gate. These are just two of the configurations explored by the authors of the Block-Recurrent Transformer. The one we describe in this section, a fixed gate in a skip configuration, is the one that obtained the best results in the original paper, and the one we use in the model we describe in the next section.

In both types of cells, learned state IDs are added to the current state vectors before they undergo any other computations. The rational is as follows: As all state vectors undergo the same computations, the resulting values would be identical, if the state vectors were indistinguishable. The state IDs allow the model to distinguish between state vectors and are implemented as simple learned positional embeddings.

As global positional embeddings don't work well for long sequences, the original paper suggests using T5-style relative position biases \cite{t5} for the self-attention module of the input token embeddings in the vertical cell. The implementation we use leverages Extrapolatable Position Embeddings \cite{xpos} instead. This adds a relative bias to the keys and queries matrices ($K$, $Q$) of the self-attention mechanism. In the next section we use the function $\text{MHR}()$ to refer to the Multi-Head Attention function, where the matrices $K$ and $Q$ of the individual attention heads undergo this procedure.

\fi

\subsubsection{{Model Details}}

\hl{After constructing the matrix $Z$, the model feeds it into a Block-Recurrent Transformer consisting of one horizontal layer and one vertical layer. The matrix $Z$ is divided into segments of size $S$ and further into blocks of size $B$. In the following we assume $|Z| < S$ and outline the process undergone by $Z$ in this case. The general case is described after.}

\hl{The blocks of $Z$ are fed into the layers of the Block-Recurrent Transformer one by one, starting from the top of $Z$. Let $Z_b$ be the $b$-th block in $Z$. After the predecessors of $Z_b$ have been processed, $Z_b$ is fed into the \textbf{horizontal cell} along with the current state vectors $C$. Here, the state vectors are first normalized and graced with learned positional embeddings as follows: }

\begin{equation}
    \overline{C} = \overline{C} + W_p,
\end{equation}

\hl{where $W_p$ is a trainable parameter. }

\hl{Next, we extract the keys and values matrices ($K_b$, $V_b$) of $Z_b$ as follows: }

\begin{equation}
    K_{b} = \overline{Z_{b}} W_K^{Z_{b}} \text{ and } V_{b} = \overline{Z_{b}} W_V^{Z_{b}},
\end{equation}

\hl{where $W_K^{Z_b} \in \mathbb{R}^{8d \times I d_k}$, and $W_V^{Z_b} \in \mathbb{R}^{8d \times I d_v}$ are trainable parameters. These matrices are stored for later use in the vertical layer as well. }

\hl{The computation proceeds as follows:}

\begin{equation}
    A_h^{(1)} = \text{MH} \left( \overline{C} W_{Q}^{(1)}, \overline{C} W_K^{C}, \overline{C} W_V^{C} \right),
\end{equation}
\begin{equation}
    A_h^{(2)} = \text{MH} \left( \overline{C} W_{Q}^{(2)}, [K_{b-1} ; K_b], [V_{b - 1} ; V_b] \right),
\end{equation}

\hl{where $W_{Q}^{(1)} \in \mathbb{R}^{8d \times I d_k}$, $W_K^{C} \in \mathbb{R}^{8d \times I d_k}$, $W_V^{C} \in \mathbb{R}^{8d \times I d_v}$, and $W_{Q}^{(2)} \in \mathbb{R}^{8d \times I d_k}$ are trainable parameters, $I$ denotes the number of attention heads, and $[* ; *]$ refers to vertical concatenation. If $b = 1$, the concatenation is skipped. $A_h^{(1)}$ is the result of self-attention on the current state vector, while $A_h^{(2)}$ results from cross-attending the state vectors with the input $Z_b$. Note, that the function MH refers to the Multi-Head Attention module as described in Section \mbox{\ref{sec:mh}}.}

\hl{The matrices $A_h^{(1)}$ and $A_h^{(2)}$ are then concatenated horizontally and projected to a suitable dimension as follows:}

\begin{equation}
    P_h = (A^{(1)} \| A^{(2)})W_h + b_h,
\end{equation}

\hl{where $W_h \in \mathbb{R}^{2 I d_v \times 8d}$ and $b_h \in \mathbb{R}^{8d}$ are trainable parameters. Subsequently, $P_h$ enters a gating mechanism along with the current state vectors. There, each state vector is multiplied element-wise with the sigmoid of a trainable vector $b_g \in \mathbb{R}^{8d}$, while the rows of $P_h$ are multiplied element-wise with the vector $1 - \sigma(b_g)$. These are then added together as shown below.}

\begin{equation}
    N = C \odot \sigma (b_g) + P_h \odot (1 - \sigma (b_g)),
\end{equation}

\hl{where $\odot$ denotes element-wise multiplication, and $\sigma$ denotes the sigmoid function. $N$ is now referred to as the next state vectors and is fed, along with $Z_b$ to a vertical cell.}

\hl{In the \textbf{vertical cell}, the state vectors are first normalized and graced with learned positional embeddings as before: }

\begin{equation}
    \overline{N} = \overline{N} + W_p.
\end{equation}

\hl{Then, they are processed as follows:}

\begin{equation}
    A^{(3)} = \text{MHR} \left( \overline{Z_b} W_{Q}^{(3)}, [K_{b - 1} ; K_b], [V_{b - 1} ; V_b] \right),
\end{equation}
\begin{equation}
    A^{(4)} = \text{MH} \left( \overline{Z_b} W_{Q}^{(4)}, \overline{N} W_K^{C}, \overline{N} W_V^{C} \right),
\end{equation}

\hl{where $W_{Q}^{(3)} \in \mathbb{R}^{8d \times I d_k}$, and $W_{Q}^{(4)} \in \mathbb{R}^{8d \times I d_k}$ are trainable parameters, and $W_K^{C}$ and $W_V^{C}$ are the same parameters used to project the state vectors in the horizontal layer. Unlike before, here, the input $Z_b$ undergoes self-attention, while cross-attending with the state vectors extracted from the previous cell $N$. Note, that the function MHR refers to a Multi-Head Attention module with Extrapolatable Position Embeddings \cite{xpos}.}

\hl{As before, the matrices $A^{(3)}$ and $A^{(4)}$ are concatenated horizontally and projected to a suitable dimension:}

\begin{equation}
    P_v = (A^{(3)} \| A^{(4)})W_v + b_v,
\end{equation}

\hl{where $W_v \in \mathbb{R}^{2 I d_v \times 8d}$ and $b_v \in \mathbb{R}^{8d}$ are trainable parameters. $Z_b$ is then added to $P_v$ over a residual connection. The resulting matrix is fed into an $FFN_{GEGLU}$ \mbox{\cite{geglu}} and added to its output over a residual connection:}

\begin{equation}
    O_b = \text{FFN}_{GEGLU}(P_v + Z_b) + P_v + Z_b.
\end{equation}

\hl{$O_b$ is the output of the Block-Recurrent Transformer for the block $Z_b$.}

\hl{If $Z$ can be divided into multiple segments, then the keys and values of the last block of the first segment are cached to be utilised by the first block of the second segment, and so on.}

\hl{The outputs of all blocks $O_1, \ldots, O_B$ are then concatenated vertically:}

\begin{equation}
    H = [O_1; \ldots; O_B] \in \mathbb{R}^{\max \{l_u, l_v\} \times 8d}.
\end{equation}

\paragraph{{Average Pooling:}} \hl{The $d_{out}$-dimensional node representations of $u$ and $v$ are finally extracted from $H$:}

\begin{equation}
    h_u = \text{Mean} \left( H [:, :4d] \right)W_{out} + b_{out} \in \mathbb{R}^{d_{out}},
\end{equation}
\begin{equation}
    h_v = \text{Mean} \left( H [:, 4d:] \right)W_{out} + b_{out} \in \mathbb{R}^{d_{out}},
\end{equation}

\hl{where $W_{out} \in \mathbb{R}^{4d \times d_{out}}$ and $b_{out} \in \mathbb{R}^{d_{out}}$ are trainable parameters, and $d_{out}$ is the output dimension.}

\subsection{{Decoder}}

\hl{We inherit the decoders for the two downstream tasks, dynamic link prediction and dynamic node classification, from DyGLib. For dynamic link prediction, the decoder is a simple MLP with one ReLu-activated hidden layer. For dynamic node classification, the decoder is an MLP with two ReLu-activated hidden layers. The outputs of both decoders are just one value.}

\section{Evaluation Methodology: Additional Details}

\subsection{Evaluation Metrics}

We use two metrics: the \textbf{Average Precision (AP)} and the \textbf{Area Under the ROC (AUC)}. 

In AP, the model being evaluated outputs a probability $p$ for each query. The threshold $\tau$ is the probability at which the predictions are considered positive, i.e., for $p \geq \tau$ the query is considered positive, while for $p < \tau$ the query is considered negative. Average Precision is given by $\text{AP} = \frac{1}{N} \sum_{i = 1}^N \left( r(\tau_i) - r(\tau_{i - 1}) \right) p(\tau_i)$,
where $p( \tau_i )$ and $r( \tau_i )$ are the precision and recall values for the classifier with threshold $\tau_i$. We set $r( \tau_0 ) = 0$. We use scikit's \cite{scikit} implementation of this metric. 

\paragraph{Area Under the ROC (AUC-ROC):} The ROC curve plots recall against the false positive rate at various classification thresholds. AUC-ROC measures the area under the ROC curve, where higher scores are desirable. Intuitively, one can think of AUC-ROC as the probability that the model can distinguish a random positive sample from a random negative sample. Again, we use scikit's \cite{scikit} implementation of this metric.

\subsection{Design Details}

Our implementation is integrated into DyGLib~\cite{dygformer}. Our implementation of BRT is heavily based on Phil Wang's implementation, which can be found under the following link: \url{https://github.com/lucidrains/block-recurrent-transformer-pytorch/tree/main}.

\subsection{Model Parameters}

We split the datasets into train, val, and test by the ratio of $70\%$-$15\%$-$15\%$.  We set the mini-batch size to $100$, and use the Adam optimizer~\cite{kingma2014adam}. We use a learning rate of 0.0001, and train for 50 epochs. We employ early stopping with a patience of either 0 or 2 depending on the dataset. The node mini-batch sampling is done sequentially in order to follow the chronological order of the interactions. Other model parameters are as follows:

\begin{itemize}
    \item Dimension of aligned encoding $d$: 50
    \item Dimension of time encoding $d_T$: 100
    \item Dimension of neighbor co-occurrence encoding $d_C$: 50
    \item Dimension of output representation $d_{out}$: 172
    \item Number of BRT cells: 2
    \item Position of the BRT horizontal cell: 1
    \item Number of attention heads $I$: 4
    \item BRT block size: 16
    \item BRT segment size: 32
    \item BRT number of state vectors: 32
\end{itemize} 

\marginpar{\large\vspace{1em}\colorbox{yellow}{\textbf{ZJkb}}\\\colorbox{yellow}{\textbf{(a)}}}

\hl{Our methodology for hyperparameter selection follows DyGLib, an established library and benchmarking infrastructure for dynamic graph learning~\mbox{\cite{dygformer}}. The values used for the parameter $s_2$ (defined in~\mbox{\ref{sec:model-ho-interactions}}) were found by means of a grid search over the universe $\{0, 1\}$ for the datasets MOOC and LastFM, and over $\{0, 1, 2, 4\}$ for the dataset CanParl. These universes were chosen so as to test the model both with and without higher order structures. In the latter case, we keep the amount of considered 2-hop interactions low (in the order of magnitude of the amount of 1-hop interactions $s_1$). This allows us to consider higher order structures while avoiding adding potential noise. Finer tuning was limited by the available compute resources.}

% These universes were chosen so as to test the model both with and without higher order structures. In the latter case, we try to keep the amount of considered 2-hop interactions low (in the order of magnitude of the amount of 1-hop interactions), so as to avoid adding too much noise.

\subsection{Dataset Details}

The details of the datasets illustrated in Section~\ref{sec:eval} \hl{as well as dataset dependent parameters} are in Table~\ref{tab:datasets}.

\begin{table}[ht]
%\vspace{-1em}
 \centering
 \scriptsize
% \resizebox{\textwidth}{!}{
  \scriptsize
 \begin{tabular}{ l|ccccc }
  \toprule
  \textbf{Data} & \textbf{Interactions} & \textbf{Sequence length} & \textbf{Patch size} & \textbf{Dropout rate} & \textbf{Patience} \\
  \midrule
%  Wikipedia & 157,474 & 32 & 1 & 0.1 & 2 \\
%  Reddit & 672,447 & 64 & 2 & 0.2 & 0 \\
  MOOC & 411,749 & 256 & 8 & 0.1 & 2 \\
  LastFM & 1,293,103 & 512 & 16 & 0.1 & 0 \\
%  Enron & 125,235 & 256 & 8 & 0.0 & 2 \\
%  Social Evo. & 2,099,519 & 32 & 1 & 0.1 & 0 \\
%  UCI & 59,835 & 32 & 1 & 0.1 & 2 \\
%  Flights & 1,927,145 & 256 & 8 & 0.1 & 0 \\
  Can. Parl. & 74,478 & 2048 & 64 & 0.1 & 2 \\
%  US Legis. & 60,396 & 256 & 8 & 0.0 & 2 \\
%  UN Trade & 507,497 & 256 & 8 & 0.0 & 2 \\
%  UN Vote & 1,035,742 & 128 & 4 & 0.2 & 0 \\
%  Contact & 2,426,279 & 32 & 1 & 0.0 & 0 \\
  \bottomrule
 \end{tabular}
 %}
 \caption{Overview of model configurations over various datasets.}
 \label{tab:datasets}
\end{table}

\end{document}